\documentclass[sigconf]{acmart}
\usepackage{algorithm}
\usepackage{algorithmic}
\AtBeginDocument{%
  }

\setcopyright{acmlicensed}
\copyrightyear{2026}
\acmYear{2026}
\acmDOI{XXXXXXX.XXXXXXX}
\acmConference[MM '26]{the 34th ACM International Conference on Multimedia}{November 10--14, 2026}{Rio de Janeiro, Brazil}
\acmSubmissionID{3022}   %%
\acmISBN{978-1-4503-XXXX-X/2026/06}

\begin{document}

\title{Rare Concept Generation via Counterfactual Inference in Diffusion Models}


\author{Zhengyuan Jiang}
\authornote{Both authors contributed equally to this research.}
\affiliation{%
	\institution{School of Computer Science and Information Engineering, Hefei University of Technology}
	\city{Hefei}
	\country{China}
}
\email{2024110489@mail.hfut.edu.cn}

\author{Haipeng Liu}
\authornotemark[1]
\affiliation{%
	\institution{School of Computer Science and Information Engineering, Hefei University of Technology}
	\city{Hefei}
	\country{China}
}
\email{hpliu\_hfut@hotmail.com}

\author{Meng Wang}
\affiliation{%
	\institution{School of Computer Science and Information Engineering, Hefei University of Technology}
	\city{Hefei}
	\country{China}
}
\email{eric.mengwang@gmail.com}

\author{Yang Wang}
\authornote{Yang Wang is the corresponding author.}
\affiliation{%
	\institution{School of Computer Science and Information Engineering, Hefei University of Technology}
	\city{Hefei}
	\country{China}
}
\email{yangwang@hfut.edu.cn}

\renewcommand{\shortauthors}{Trovato et al.}

\begin{abstract}
Rare concept generation focuses on synthesizing customized images conditioned on text prompts that describe objects with unusual attributes. Previous works failed to align the generated images with rare concepts, resulting in incorrect attribute rendering or inconsistent composition of concepts. Such failures, as we observed, stem from the inherent common knowledge bias in the training stage of diffusion models, where objects are strongly associated with their common attributes, making it difficult to break these associations when generating rare concepts. To address such challenges, in this paper, we propose a novel \textbf{C}ounterfactual \textbf{I}nference-based \textbf{Diff}usion approach, dubbed \textbf{CI-Diff}. \textbf{CI-Diff} blocks the interference of the model's inherent common knowledge bias and utilizes the Natural Direct Effect to capture the independent influence of the text prompt of rare concepts on image generation so that decoupling the unusual attributes from the rare concepts. To this end, we reformulate the classifier-free guidance mechanism to highlight the atypical attributes. To the best of our knowledge, we are the first to introduce causal inference into the rare concept generation task. Extensive experiments on the RareBench benchmark validate the superiority of \textbf{CI-Diff} over state-of-the-art diffusion models. Our code can be accessed from \url{https://github.com/200204jzy/CI-Diff}.

\end{abstract}

%\begin{CCSXML}
%<ccs2012>
%  <concept_id>00000000.0000000.0000000</concept_id>
%  <concept_desc>Do Not Use This Code, Generate the Correct Terms for Your Paper</concept_desc>
%  <concept_significance>500</concept_significance>
%  <concept_id>00000000.00000000.00000000</concept_id>
%  <concept_desc>Do Not Use This Code, Generate the Correct Terms for Your Paper</concept_desc>
%  <concept_significance>300</concept_significance>
%  <concept_id>00000000.00000000.00000000</concept_id>
%  <concept_desc>Do Not Use This Code, Generate the Correct Terms for Your Paper</concept_desc>
%  <concept_significance>100</concept_significance>
%  <concept_id>00000000.00000000.00000000</concept_id>
%  <concept_desc>Do Not Use This Code, Generate the Correct Terms for Your Paper</concept_desc>
%  <concept_significance>100</concept_significance>
%</ccs2012>
%\end{CCSXML}



%\ccsdesc[500]{Do Not Use This Code~Generate the Correct Terms for Your Paper}
%\ccsdesc[300]{Do Not Use This Code~Generate the Correct Terms for Your Paper}
%\ccsdesc{Do Not Use This Code~Generate the Correct Terms for Your Paper}
%\ccsdesc[100]{Do Not Use This Code~Generate the Correct Terms for Your Paper}

%\keywords{Do, Not, Use, This, Code, Put, the, Correct, Terms, for,
%  Your, Paper}
\begin{teaserfigure}
	\centering
	\includegraphics[width=0.9\textwidth]{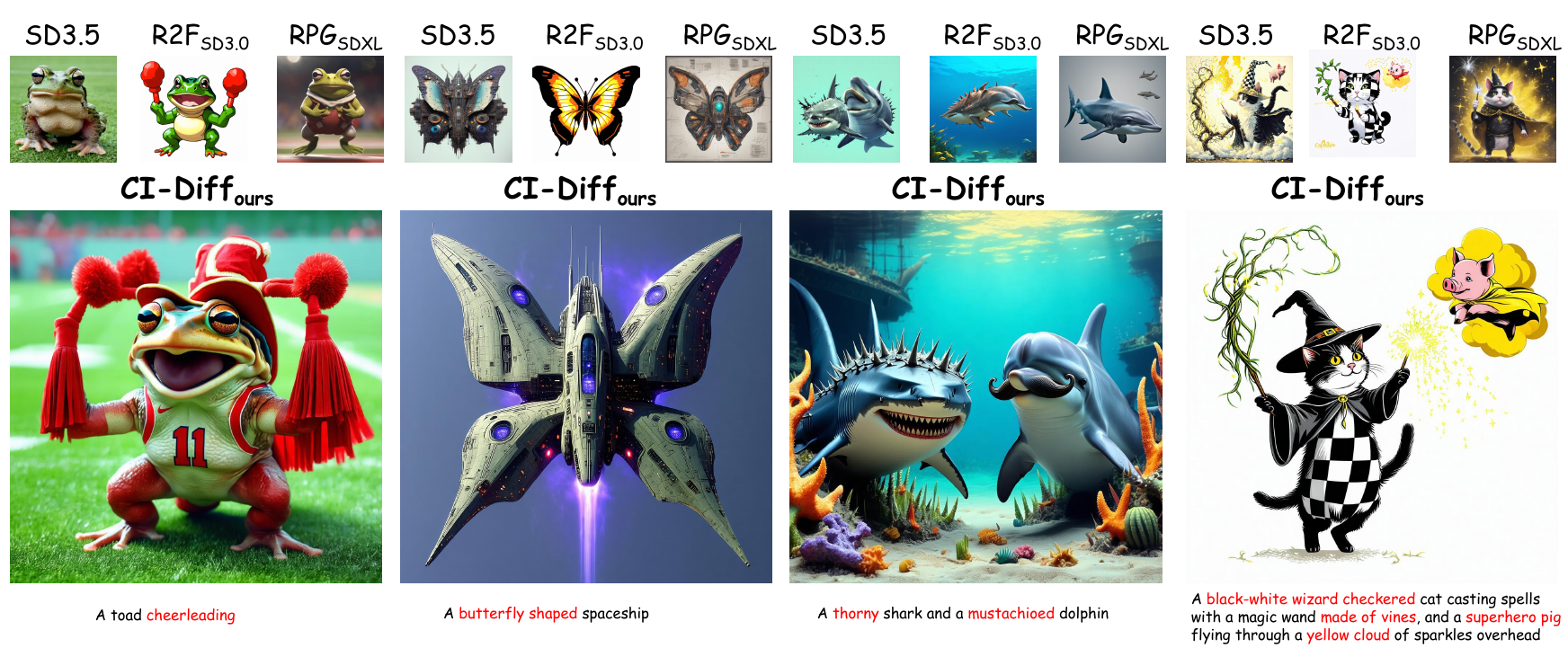} % 替换为你的文件名
	\caption{Generated images from rare concepts (unusual attributes are highlighted in red). In contrast to the unsatisfactory performance of other methods, our proposed approach, CI-Diff, yields significantly better results.}
	\label{fig:teaser}
\end{teaserfigure}

%%\received{20 February 2007}
%%\received[revised]{12 March 2009}
%%\received[accepted]{5 June 2009}

\maketitle

\section{Introduction}
\label{sec:intro}

Recent advancements in text-to-image (T2I) diffusion models have achieved unprecedented success in generating highly realistic and diverse images \cite{diffusion_sucess, diffusion_sucess_1, diffusion_sucess_2, diffusion_sucess_3, diffusion_sucess_4, shengcheng}. With the rapid development of artificial intelligence \cite{dfq1, dfq2, dfq3, dfq4, shibie, fengge}, users increasingly push the boundary by exploring rare or highly imaginative prompts \cite{shixiong1, shixiong2,shixiong3, fenge, fenge1}. Rare concept generation refers to the synthesis of customized images based on text prompts that describe objects with unusual attributes \cite{R2F}.

\begin{figure}[t]
	\centering
	\includegraphics[width=0.7\linewidth]{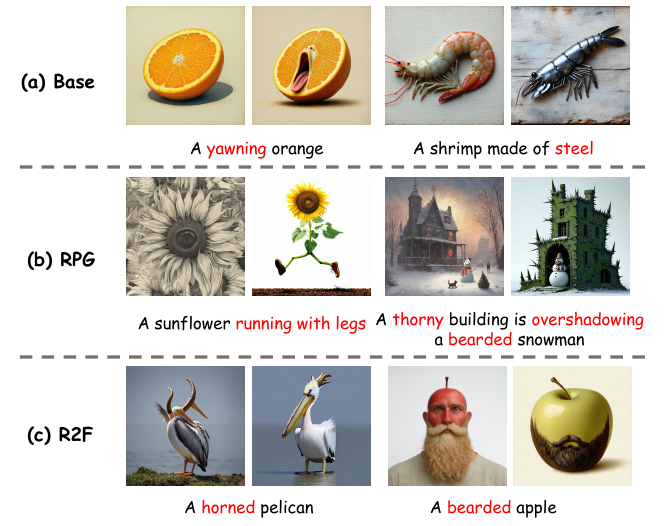}
	\caption{Weaknesses of existing methods. Conventional generation approaches are restrained by common knowledge bias and struggle to produce rare-concept images faithfully, whereas our \textbf{CI-Diff} achieves high-quality rare concept generation. The right image in each group shows results generated by our \textbf{CI-Diff}.}
	\label{fig:weak}
	\Description{Comparison of rare concept generation results for Base, RPG, and R2F methods.}
\end{figure}

With the continuous iteration of architectures and strategies \cite{diff_survey, diff_survey_2, mir}, from the early Stable Diffusion 1.5 \cite{LDM_sd15} to SDXL \cite{sdxl}, and further to the latest SD 3.0 \cite{SD3035}, the image-text alignment capability of such pre-trained models in common text scenarios has been increasingly enhanced. However, when faced with rare concepts such as ``a toad cheerleading'' in Fig.~\ref{fig:teaser}, the model struggles to accurately represent their unusual attributes. The reason lies in the fact that such models are all trained on massive conventional datasets\cite{R2F}, where rare concepts appear with extremely low frequency, leading the model to gradually form a significant \textbf{Common Knowledge Bias $K$} during training. In simple terms, the model strongly binds specific concepts to common attributes, making it difficult to overcome its inherent biases and express unusual attributes. As shown in Fig.~\ref{fig:weak}(a), when generating images corresponding to ``A yawning orange'' and ``A shrimp made of steel'', the model outputs still lean toward common concepts, and the unusual attributes in the text prompts fail to be effectively expressed.

To accurately render unusual attributes, recent works attempt to leverage large language models (LLMs) to assist image generation for improving the semantic alignment of text prompts. These methods \cite{llm1, llm2, llm3, llm4, llm7} decompose input text prompts into different sub-prompts via LLMs and extract corresponding bounding boxes. On this basis, RPG \cite{RPG} further provides detailed descriptions for each sub-prompt through a recaption mechanism, which is to submit each sub-prompt to the LLM and conduct a detailed description without altering the core meaning of the sub-prompt, and divides image generation regions to assign matched sub-prompts, decomposing complex combinatorial generation tasks into parallel local generation tasks. Finally, RPG \cite{RPG} ensures the semantic consistency between the generated images and the target texts through closed-loop editing. Nevertheless, RPG \cite{RPG} still has significant limitations: 1) Some rare concepts, even after being rewritten by LLMs, can hardly be transformed into common concepts that the model can generate perfectly. As shown in Fig.~\ref{fig:weak}(b), for the rare concept ``a sunflower with legs running'', even with elaborate rewritten descriptions provided by LLMs, the diffusion model still struggles to break the strong semantic binding between ``plant'' and ``static'', and fails to effectively decouple the unusual attribute of ``running''. 2) RPG \cite{RPG} struggles prompts involving overlapping entities and complex spatial relations. For example, for ``A thorny building overshadowing a bearded snowman'', conflicts between LLM spatial planning and internal diffusion attention prevent accurate spatial and morphological rendering.

Different from RPG \cite{RPG}, R2F \cite{R2F} leverages the rich semantic knowledge of large language models (LLMs) to first identify unusual attributes (e.g., "bearded") from rare concepts like "a bearded apple", and matches them with semantically consistent common concepts such as "a bearded person". Subsequently, it designs a progressive guidance strategy from "rare concepts to common concepts" in the diffusion sampling stage for image generation. Nevertheless, this method still suffers from an obvious weakness. The common concepts introduced by LLMs often contain a large amount of redundant semantic information irrelevant to the target unusual attributes (such as human characteristics), which is highly likely to cause additional interference with the generation process during the denoising stage. As shown in Fig.~\ref{fig:weak}(c), when given prompts such as "A horned pelican" and "A bearded apple", although the generated results can reflect the unusual attributes, the shapes of the objects are significantly distorted. To sum up, both RPG \cite{RPG} and R2F \cite{R2F} attempt to assist diffusion models in rare concept generation by introducing external semantic knowledge from large language models to restructure and decompose input text prompts, which not only ignores the potential of rare concept generation inherent in pre-trained models, but also fails to fundamentally decouple unusual attributes from rare concepts.

\begin{figure*}[h]
	\centering
	\includegraphics[width=0.8\linewidth]{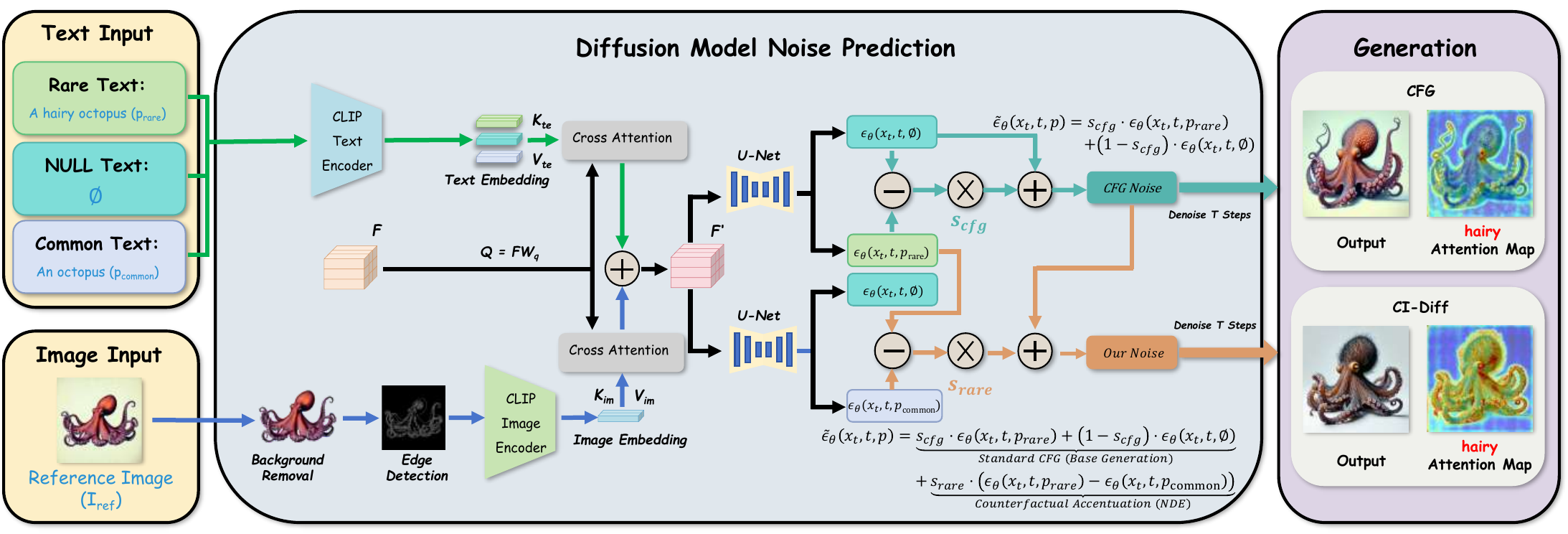}
	\caption{Overview of our proposed \textbf{CI-Diff} framework. The framework adopts a dual-path text input design, which takes the rare prompt text $P_{rare}$ and the corresponding common subject text $P_{common}$ as inputs, and generates reference images matching target objects based on $P_{common}$. Leveraging counterfactual causal inference, our method decouples and amplifies the unusual attributes contained in rare texts at the noise prediction layer, mitigating the suppression of special feature rendering caused by the model’s inherent common knowledge bias. Meanwhile, edge contour maps extracted from reference images are introduced as additional constraint information during the iterative diffusion denoising process, so as to guarantee the complete morphology and structural stability of the main objects in generated images.}
	\Description{Three subfigures showing ablation studies.}
	\label{fig:kuangjiatu}
\end{figure*}

Based on the above, we propose a novel training-free approach, namely the \textbf{C}ounterfactual \textbf{I}nference-based \textbf{Diff}usion approach (\textbf{CI-Diff}). To the best of our knowledge, we are the first to introduce causal inference into the rare concept generation task. Specifically, we first construct a real causal graph of text-to-image generation via causal inference, and clarify how the internal \textbf{Common Knowledge Bias ($K$)} interferes with the generation process. Then, by constructing counterfactual scenarios, we observe the Natural Direct Effect (NDE) of text prompts on image generation, which can the eliminate the interference of \textbf{Common Knowledge Bias} and captures the independent impact of text on image generation, thus improving the text-image alignment accuracy. Finally, we achieve the decoupling of unusual attributes from rare concepts by comparing the NDEs of rare concepts and common concepts.Technically, we parameterize the Natural Direct Effect based on Classifier-Free Guidance (CFG) \cite{CFG}, thereby mapping the causal effect of unusual attributes to the noise space for practical implementation. In addition, we propose the \textbf{T}emporal \textbf{M}orphological \textbf{F}idelity \textbf{A}nchoring Strategy \textbf{(TMFA)}, which injects the background-removed edge images within specific time steps during the denoising process to ensure the shape consistency of the object.

In summary, our major contributions are summarized below: 
{
\leftmargini=5mm
\begin{itemize}
	\item We propose \textbf{CI-Diff}, a rare concept generation approach based on counterfactual inference, which introduces causal inference into text-to-image generation for the first time. By constructing causal graphs and counterfactual scenarios to mine the Natural Direct Effect, our method blocks the interference of the internal \textbf{Common Knowledge Bias} and accurately extracts the independent impact of text prompts on image generation.
	\item We design a decoupling scheme by comparing the Natural Direct Effects between common and rare concepts. With Classifier-Free Guidance, we map the causal effects of unusual attributes to the noise space, achieving effective decoupling and enhanced expression of unusual attributes in a training-free manner.
	\item We present the Temporal Morphological Fidelity Anchoring Strategy, which injects background-removed edge images within specific time steps during diffusion denoising. This strategy strengthens the representation of unusual attributes while effectively maintaining the shape consistency of the objects throughout the generation process.
\end{itemize}
}
\noindent The extensive experimental results validate the advantages of our \textbf{CI-Diff} against other models for rare concept generation.

\section{Method}
\label{sec:method}

In this section, we technically explain our pipeline shown in Fig.~\ref{fig:kuangjiatu} in more details. Sec.~\ref{sec:preliminaries} introduces the preliminaries regarding diffusion models and causal inference. Secs.~\ref{sec:counterfactual_reasoning} and~\ref{sec:concept_guidance} elaborate on how to decouple unusual attributes via counterfactual inference and the corresponding implementation within the diffusion process. We finally propose a novel Temporal Morphological Fidelity Anchoring (TMFA) strategy to ensure the consistency of the object shape during generation in Sec.~\ref{sec:tmfa}.

\subsection{Preliminaries}
\label{sec:preliminaries}

\subsubsection{Diffusion Model}
Given a text prompt $p$ that contains rare concepts, the goal of rare concept generation is to generate the corresponding image $I_{\text{gen}} \in \mathbb{R}^{3 \times H \times W}$, such that the generated content accurately presents unusual attributes and maintains high visual consistency with the text prompt. To achieve the alignment between complex text prompt and generated images, state-of-the-art methods \cite{t2i1, t2i2, t2i3, t2i4, t2i5} predominantly build upon pre-trained text-to-image latent diffusion models, primarily utilizing Stable Diffusion (SD) \cite{LDM_sd15} as their foundation. Within this architecture, a Variational Auto-Encoder (VAE) \cite{VAE} encodes the pixel-space image $I \in \mathbb{R}^{3 \times H \times W}$ into a latent representation $z \in \mathbb{R}^{4 \times \frac{H}{8} \times \frac{W}{8}}$, significantly reducing computational complexity without compromising visual quality. The model then performs
the forward diffusion process, denoising process and Classifier-Free Guidance sampling process in the latent space:

\noindent \textbf{Forward Diffusion Process.} Consisting of $T$ timesteps, the forward process in Stable Diffusion builds upon the standard Denoising Diffusion Probabilistic Model (DDPM) \cite{DDPM}. Specifically, it progressively injects Gaussian noise $\epsilon \sim \mathcal{N}(0, I)$ to corrupt a clean input $z_0$ into a noisy state $z_T$. Thus, at any given timestep $t \in [0, T]$, the intermediate state $z_t$ is formulated as:
\begin{equation}
	z_t(z_0, \epsilon) = \sqrt{\bar{\alpha}_t} z_0 + \sqrt{1 - \bar{\alpha}_t} \epsilon,
	\label{eq:diffusion_forward}
\end{equation}
where $\bar{\alpha}_t$ denotes the corresponding noise level.

\noindent \textbf{Reverse Denoising Process.} In the reverse phase, Stable Diffusion learns to predict the injected noise within the noisy sample, guided by the condition $c$, iteratively denoising it across $T$ timesteps. Starting from pure noise $z_T$ drawn from a standard Gaussian distribution, the optimization objective for the UNet-based \cite{unet} denoising model $\epsilon_\theta$ at timestep $t$ is defined as:
\begin{equation}
	\mathcal{L}_{\text{LDM}} = \mathbb{E}_{z_0, c, t \sim \mathcal{U}(0, T), \epsilon \sim \mathcal{N}(0, \mathbf{I})} \left\| \epsilon_\theta(z_t, t, \tau_\theta(c)) - \epsilon \right\|_2^2,
	\label{eq:ldm_loss}
\end{equation}
where $\tau_\theta(\cdot)$ is the CLIP \cite{clip} encoder, and $\left\| \cdot \right\|_2$ denotes the $\ell_2$ norm.

\noindent \textbf{Classifier-Free Guidance.} In the inference sampling process, Stable Diffusion adopts Classifier-Free Guidance (CFG) \cite{CFG} to improve the semantic alignment between generated images and the given text prompt. Starting from the noisy sample $z_T$, the model performs both the prediction based on the condition $c$ and the unconditional  prediction based on empty conditioned $\emptyset$ at each timestep $t$, and computes the final noise prediction $\epsilon_\theta$ as below:
\begin{equation}
	\hat{\epsilon}_\theta(z_t, t, c) = \omega \epsilon_\theta(z_t, t, c) + (1 - \omega) \epsilon_\theta(z_t, t, \emptyset),
	\label{eq:cfg_guidance}
\end{equation}
where $\omega$ denotes the guidance scale. After T timesteps, the
denoised latent representation $z_0'$ is reconstructed into pixel space via the decoder $\mathcal{D}$ to generate the final result $I_{\text{gen}}$.

\subsubsection{Counterfactual Causal Inference}
\textbf{Causal inference} \cite{yingguo_1, yingguo_2, yingguo_3, yingguo_4, VQA} refers to a mathematical framework that characterizes causal structures using a causal graph and evaluates causal effects between variables in complex systems through counterfactual intervention. For a causal graph consisting of three variables $P$, $K$, and $I$, if the variable $P$ has a direct effect on the variable $I$, we say that $P$ is the child of $I$, i.e., $P \rightarrow I$. If $P$ has an indirect effect on $I$ via the variable $K$ , we say that $K$ acts as a mediator between $P$ and $I$, i.e., $P \rightarrow K \rightarrow I$. With counterfactual notation, the aforementioned causal graph can be transformed into the following formulas:
\begin{equation}
	I_{p,k} = I(P=p, K=k).
\end{equation}

In the factual scenario, the mediator takes the value  $k = K_p = K(P = p)$. In the corresponding counterfactual scenario, different values are assigned to $P$ when separately calculating $K$ and $I$.  For instance, $I_{p^\ast,K_{p}}$ denotes a state where $P$ is intervened to be $p^\ast$, yet $K$ retains the value it would have naturally taken had $P$ been $p$, i.e., $I_{p^\ast,K_{p}} = I(P=p^\ast, K=K(P=p))$.

\noindent \textbf{Causal Effect} \cite{yingguo_5, yingguo_7} refers to the comparison of potential outcomes for the same individual under distinct treatment regimes. Let $P=p$ denote the ``treatment condition'' and $P=p^\ast$ denote the ``no-treatment condition.'' The total effect (TE) of treatment $P=p$ on variable $I$ evaluates the discrepancy between these two hypothetical scenarios, which is formulated as:
\begin{equation}
	\text{TE} = I_{p, K_p} - I_{p^\ast, K_{p^\ast}}.
	\label{eq:total_effect}
\end{equation}

The total effect can be decomposed into the natural direct effect (NDE) and the total indirect effect (TIE). NDE denotes the effect of $P$ on $I$ with the mediator $K$ blocked, and expresses the increase in $I$ with $P$ changing from $p^\ast$ to $p$, while $K$ is set to the value it would have obtained at $P=p^\ast$, meaning that the response of $K$ to the treatment $P=p$ is disabled:
 \begin{equation}
 	\text{NDE} = I_{p, K_{p^\ast}} - I_{p^\ast, K_{p^\ast}}.
 	\label{eq:nde}
 \end{equation}
 
TIE is the difference between TE and NDE, denoted as:
\begin{equation}
	\text{TIE} = TE - NDE = I_{p,K_p} - I_{p,K_{p^\ast}}.
	\label{eq:tie}
\end{equation}

\begin{figure}[t]
	\centering
	\includegraphics[width=0.85\linewidth]{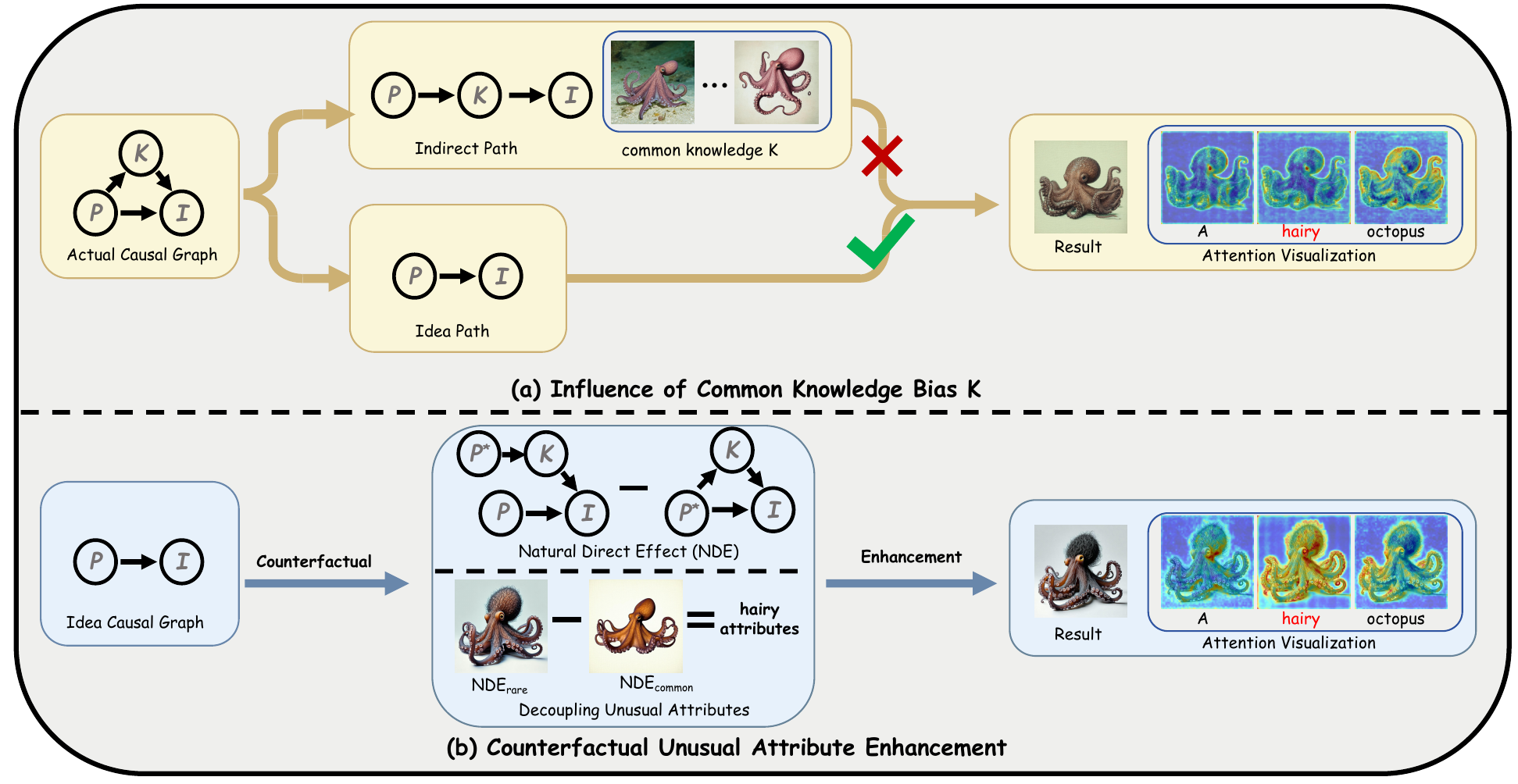}
	\caption{Comparison between conventional generation and the proposed counterfactual generation. Conventional pipelines are dominated by common knowledge bias and fail to render rare attributes, our counterfactual method eliminates such bias interference to highlight unusual features.}
	\label{fig:yinguotu}
	\Description{}
\end{figure}

\subsection{Cause-Effect Look at Rare Concept Generation}
\label{sec:counterfactual_reasoning}

The ideal rare concept generation task aims to generate a corresponding image $I$ based on a rare text prompt $P$, i.e., $P \rightarrow I$. However, as discussed in Sec.~\ref{sec:intro}, pre-trained models are susceptible to interference from \textbf{common knowledge bias $K$}, which forms a $P \rightarrow K \rightarrow I$ path, leading to the suppression of unusual attributes. To address this, we propose a counterfactual approach, which estimates the causal effect of text prompts on images by blocking the mediating effect of $K$. 

The counterfactual scenario is defined as follows: the text prompt $P$ is set to rare text prompt $p_{rare}$ (e.g., ``a hairy octopus''), while $K$ is set to the value it would attain when the text prompt is $p^*$ (where $p^*$ denotes the no treatment condition, i.e., no text prompt is provided). Since the response of the mediator $K$ to the input is blocked, the model is compelled to rely solely on the text prompt for image generation. Fig.~\ref{fig:yinguotu} illustrates the comparison between conventional generation and counterfactual generation. By comparing the counterfactual outcome with the ``no treatment'' baseline, we can derive the Natural Direct Effect ($NDE$) of $p_{rare}$ on the image $I$:
\begin{equation}
	\text{NDE}_{\text{rare}} = I_{p_{\text{rare}}, K_{p_{\text{rare}}^*}} - I_{p_{\text{rare}}^*, K_{p_{\text{rare}}^*}}.
	\label{eq:nde_rare}
\end{equation}

Since the effect of $P$ on $K$ is blocked, $NDE$ explicitly captures the direct influence of the text prompt on the generation. Furthermore, if the input text is replaced with a common text prompt $p_{common}$ (e.g., ``an octopus'') by stripping away the unusual modifiers, its Natural Direct Effect on image $I$ is expressed as:
\begin{equation}
	\text{NDE}_{\text{common}} = I_{p_{\text{common}}, K_{p_{\text{common}}^*}} - I_{p_{\text{common}}^*, K_{p_{\text{common}}^*}}.
	\label{eq:nde_normal}
\end{equation}
By directly subtracting $NDE_{common}$ from $NDE_{rare}$, we can effectively decouple the direct effect of the unusual attribute (e.g., ``hairy'') on the generated image:
\begin{equation}
	\text{NDE}_{\text{pure\_rare}} = \text{NDE}_{\text{rare}} - \text{NDE}_{\text{common}}.
	\label{eq:nde_pure_rare}
\end{equation}

\noindent \textbf{Intuition 1: }\textit{We derive the direct effects of both rare and common concepts on the image and subsequently calculate the difference between them as formulated in Eq.~(\ref{eq:nde_pure_rare}). As illustrated in Fig.~\ref{fig:yinguotu}, the results generated by Eq.~(\ref{eq:nde_pure_rare}) demonstrate that such a counterfactual intervention can effectively decouple the unusual attributes. For example, in the attention visualization shown in Fig.~\ref{fig:yinguotu}(b), the attention maps derived from the text ``hairy'' are significantly concentrated on the object's surface, leading to the successful generation of a ``hairy octopus''. }

We perform inference by enhancing the Natural Direct Effect ($NDE$) of unusual attributes, which is fundamentally distinct from previous methods such as R2F \cite{R2F} and RPG \cite{RPG}.

\subsection{Implementation}
\label{sec:concept_guidance}
As detailed in Sec.~\ref{sec:counterfactual_reasoning}, the core idea of $NDE$ is to extract the direct influence of text prompts on image generation by comparing the differences in outputs under the ``with text prompt'' and ``without text prompt (null prompt)'' conditions. Benefiting from the random condition dropout strategy employed by diffusion models during the pre-training stage, a single model can simultaneously model both conditional and unconditional noise predictions. During the inference stage, the Classifier-Free Guidance (CFG) mechanism is typically utilized to perform linear extrapolation on the difference between conditional and unconditional noise predictions, thereby improving image-text alignment, as shown in Eq.~(\ref{eq:cfg_guidance}). Based on this, we rewrite Eqs.~(\ref{eq:nde_rare}) and~(\ref{eq:nde_normal}) into the CFG format:
\begin{equation}
	\text{NDE}_{\text{rare}} = s \cdot \epsilon_\theta(x_t, t, p_{\text{rare}}) + (1 - s) \epsilon_\theta(x_t, t, \emptyset).
	\label{eq:nde_noise_rare}
\end{equation}
\begin{equation}
	\text{NDE}_{\text{common}} = s \cdot \epsilon_\theta(x_t, t, p_{\text{common}}) + (1 - s) \epsilon_\theta(x_t, t, \emptyset).
	\label{eq:nde_noise_normal}
\end{equation}

By substituting the above two equations into Eq.~(\ref{eq:nde_pure_rare}), we can further derive the noise expression for the unusual attributes:
\begin{equation}
	\text{NDE}_{\text{pure\_rare}} = s \cdot \left( \epsilon_\theta(x_t, t, p_{\text{rare}}) - \epsilon_\theta(x_t, t, p_{\text{common}}) \right).
	\label{eq:nde_noise_pure}
\end{equation}

In the actual generation process, relying solely on Eq.~(\ref{eq:nde_noise_pure}) for image generation may lead to the loss of the overall image structure, as this term is primarily used to capture the semantic information of unusual attributes. To address this issue, we superimpose the ``counterfactual accentuation term'' extracted by NDE as a refined compensation onto the standard Classifier-Free Guidance (CFG). The final predicted value $\tilde{\epsilon}_\theta$ of the denoising network is formulated as follows:
\begin{equation}
	\begin{aligned}
		\tilde{\epsilon}_\theta(x_t, t, c) &= \underbrace{s_{\text{cfg}} \cdot \epsilon_\theta(x_t, t, p_{\text{rare}}) + (1 - s_{\text{cfg}}) \cdot \epsilon_\theta(x_t, t, \emptyset)}_{\text{Standard CFG (Base Generation)}} \\
		&\quad + \underbrace{s_{\text{rare}} \cdot \left( \epsilon_\theta(x_t, t, p_{\text{rare}}) - \epsilon_\theta(x_t, t, p_{\text{common}}) \right)}_{\text{Counterfactual Accentuation (NDE)}}
	\end{aligned},
	\label{eq:final_guidance}
\end{equation}
where $s_{\text{cfg}}$ is the base guidance scale used to maintain text consistency and image quality, and $s_{\text{rare}}$ denotes the injection strength of unusual attributes, used to amplify the feature expression of unusual attributes.

\noindent \textbf{Intuition 2:} Specifically, Eq.~(\ref{eq:final_guidance}) consists of two components, striking a balance between maintaining the structural stability of the image and highlighting unusual attributes:
{
\leftmargini=4mm
\begin{itemize}
 \item\textbf{ $s_{\text{cfg}} \cdot \epsilon_\theta(x_t, t, p_{\text{rare}}) + (1 - s_{\text{cfg}}) \cdot \epsilon_\theta(x_t, t, \emptyset)$:} \textit{Utilizes the standard CFG mechanism to ensure that the generated image possesses the features of the rare concept while ensuring the quality of image generation and consistency with the base semantics.}
\item \textbf{$s_{\text{rare}} \cdot (\epsilon_\theta(x_t, t, p_{\text{rare}}) - \epsilon_\theta(x_t, t, p_{\text{common}}))$:} \textit{The counterfactual accentuation term. By comparing the Natural Direct Effect of rare text prompts and common text prompts, it accurately captures and amplifies the intervention capability of unusual attributes on the generation process.}
\end{itemize}
}
\begin{figure}[t]
	\centering
	\includegraphics[width=0.85\linewidth]{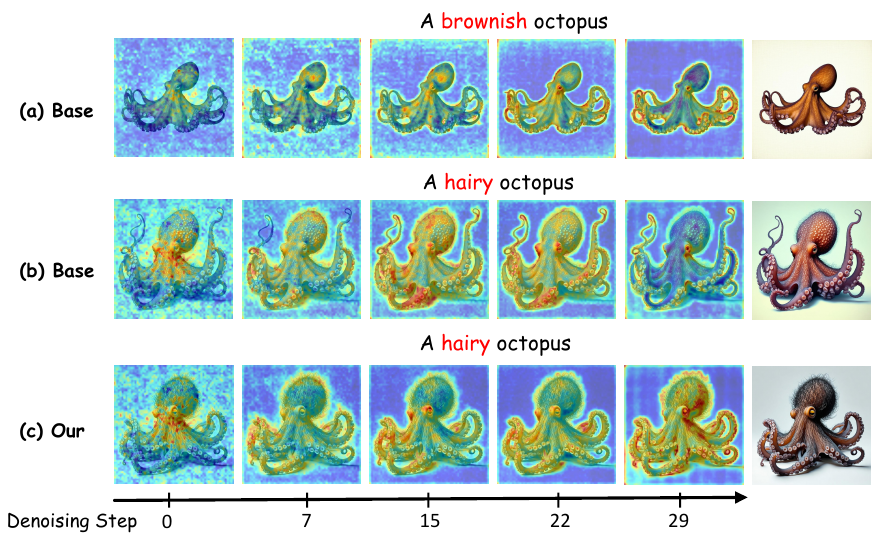}
	\caption{Evolution of attention maps for unusual attribute tokens during the denoising process. Conventional models lose focus on rare attributes in later denoising steps, while our CI-Diff stably anchors attention to unusual features.}
	\label{fig:zhuyili_step}
	\Description{Three rows of attention heatmaps across different denoising timesteps. The top row shows stable attention for a normal prompt. The middle row shows dispersing and decaying attention for a rare prompt. The bottom row shows our method correcting the rare prompt's attention to stay focused and strong over time.}
\end{figure}

To verify this strategy, Fig.~\ref{fig:zhuyili_step} illustrates the evolution of attention maps for unusual attribute tokens. As shown in Fig.~\ref{fig:zhuyili_step}(a), when the model processes common text prompts (e.g., ``A brownish octopus''), its attention stably focuses on the main object with high activation. Conversely, Fig.~\ref{fig:zhuyili_step}(b) shows that when the model processes rare text prompts, the attention diffuses and its intensity decays sharply in the middle and late stages of denoising, ultimately leading to generation failure. In contrast, as depicted in Fig.~\ref{fig:zhuyili_step}(c), when the model applies our counterfactual accentuation term, the attention for unusual attributes precisely converges onto the object and remains highly activated, thereby ensuring the successful generation of the rare concept.

\begin{table*}[h]
	\centering
	\caption{T2I alignment performance of \textbf{CI-Diff} and diffusion baselines on RareBench. Where C denotes CLIP-T score and H denotes HPSv2 score. Our CI-Diff attains optimal CLIP-T and HPSv2 score against all baseline methods.}
	\label{tab:clip_hpsv2}
	\resizebox{0.9\textwidth}{!}{
		\begin{tabular}{l|cc|cc|cc|cc|cc|cc|cc|cc|cc|cc}
			\toprule
			\textbf{Models} & \multicolumn{10}{c|}{\textbf{Single Object}} & \multicolumn{6}{c|}{\textbf{Multi Objects}} & \multicolumn{4}{c}{\textbf{Extend Objects}} \\ 
			& \multicolumn{2}{c|}{Property} & \multicolumn{2}{c|}{Shape} & \multicolumn{2}{c|}{Texture} & \multicolumn{2}{c|}{Action} & \multicolumn{2}{c|}{Complex} & \multicolumn{2}{c|}{Concat} & \multicolumn{2}{c|}{Relation} & \multicolumn{2}{c|}{Complex} & \multicolumn{2}{c|}{Style} & \multicolumn{2}{c}{Scene} \\ 
			& C & H & C & H & C & H & C & H & C & H & C & H & C & H & C & H & C & H & C & H \\ 
			\midrule
			% --- Group 1: SDXL Pair ---
			SDXL & 30.15 & 25.64 & 31.43 & 24.62 & 32.40 & 26.84 & 31.34 & 25.56 & 33.21 & 28.44 & 31.86 & 24.64 & 33.13 & 26.53 & 36.39 & 29.63 & 33.96 & 26.71 & 35.07 & 28.24 \\ 
			\textbf{Our+SDXL} & \textbf{30.43} & \textbf{26.67} & \textbf{31.60} & \textbf{26.20} & \textbf{32.41} & \textbf{27.95} & \textbf{31.53} & \textbf{27.02} & \textbf{33.60} & \textbf{29.56} & \textbf{32.32} & \textbf{25.31} & \textbf{33.69} & \textbf{26.72} & \textbf{36.61} & \textbf{30.67} & \textbf{34.60} & \textbf{28.10} & \textbf{35.75} & \textbf{28.98} \\ 
			\midrule
			% --- Group 2: RealV Pair ---
			RealVisXL & 32.14 & 28.58 & 33.06 & 26.90 & 33.73 & 29.29 & 33.49 & 27.96 & 34.60 & 30.60 & 33.23 & 28.74 & 34.85 & 28.94 & 36.93 & 32.28 & 34.13 & 29.29 & 36.35 & 29.41 \\ 
			\textbf{Our+RealVisXL} & \textbf{32.21} & \textbf{29.81} & \textbf{33.23} & \textbf{28.79} & \textbf{34.16} & \textbf{31.03} & \textbf{33.55} & \textbf{29.79} & \textbf{34.66} & \textbf{32.57} & \textbf{33.54} & \textbf{29.74} & \textbf{34.92} & \textbf{30.85} & \textbf{37.10} & \textbf{32.63} & \textbf{34.94} & \textbf{29.79} & \textbf{36.43} & \textbf{31.40} \\ 
			\midrule
			% --- Group 3: SOTA Comparison ---
			SD1.5 & 29.89 & 25.44 & 30.39 & 23.75 & 30.37 & 25.00 & 29.80 & 23.87 & 31.57 & 24.72 & 29.57 & 22.92 & 31.68 & 24.07 & 33.86 & 23.20 & 33.32 & 23.12 & 32.68 & 25.45 \\ 
			Flux & 30.60 & 30.02 & 31.36 & 27.73 & 31.92 & 30.30 & 31.69 & 29.35 & 34.18 & 31.53 & 32.53 & 30.34 & 33.79 & 29.83 & 35.57 & 32.40 & 34.72 & 30.60 & 35.52 & 31.31 \\ 
			PixArt-$\alpha$ & 29.27 & 29.66 & 31.19 & 28.03 & 32.66 & 30.57 & 30.93 & 29.39 & 33.45 & 31.94 & 30.89 & 28.46 & 33.04 & 30.29 & 35.29 & 31.97 & 34.88 & 31.05 & 34.56 & 31.91 \\ 
			SynGen & 31.14 & 25.08 & 31.38 & 23.07 & 29.32 & 24.31 & 29.94 & 23.24 & 32.88 & 25.68 & 29.73 & 22.79 & 31.25 & 22.42 & 31.87 & 22.19 & 32.67 & 24.27 & 33.13 & 26.92 \\ 
			RPG & 29.98 & 26.89 & 30.33 & 24.85 & 32.04 & 27.29 & 29.90 & 26.68 & 32.84 & 29.85 & 30.31 & 25.37 & 32.50 & 26.97 & 34.18 & 29.57 & 34.06 & 27.28 & 34.84 & 29.45 \\ 
			SD3.0 & 30.63 & 29.42 & 33.44 & 28.18 & 31.70 & 29.43 & 32.46 & 29.01 & 33.08 & 31.26 & 32.42 & 30.07 & 34.11 & 31.12 & 35.76 & 33.13 & 33.84 & 30.24 & 34.88 & 31.36 \\ 
			R2F+SD3.0 & 30.84 & 29.22 & 32.15 & 28.08 & 33.17 & 30.01 & 31.62 & 29.11 & 33.02 & 30.28 & 33.08 & 28.97 & 33.52 & 28.74 & 34.79 & 31.37 & 34.06 & 29.58 & 35.28 & 31.38 \\ 
			SD3.5 & 31.86 & 29.26 & 32.94 & 28.40 & 33.18 & 29.91 & 32.73 & 29.87 & 34.11 & 31.71 & 33.46 & 30.15 & 34.67 & 30.95 & 36.97 & 33.62 & 33.93 & 30.17 & 35.04 & 31.36 \\ 
			\textbf{Our+SD3.5} & \textbf{\textcolor{red}{32.64}} & \textbf{\textcolor{red}{30.17}} & \textbf{\textcolor{red}{33.61}} & \textbf{\textcolor{red}{29.56}} & \textbf{\textcolor{red}{34.40}} & \textbf{\textcolor{red}{31.65}} & \textbf{\textcolor{red}{34.05}} & \textbf{\textcolor{red}{30.51}} & \textbf{\textcolor{red}{34.76}} & \textbf{\textcolor{red}{32.69}} & \textbf{\textcolor{red}{34.44}} & \textbf{\textcolor{red}{31.09}} & \textbf{\textcolor{red}{35.43}} & \textbf{\textcolor{red}{31.75}} & \textbf{\textcolor{red}{37.77}} & \textbf{\textcolor{red}{34.17}} & \textbf{\textcolor{red}{35.53}} & \textbf{\textcolor{red}{31.17}} & \textbf{\textcolor{red}{36.46}} & \textbf{\textcolor{red}{31.98}} \\ 
			\bottomrule 
		\end{tabular} 
	} 
\end{table*}

\begin{table*}[t]
	\centering
	\caption{Generation quality of unusual attributes of \textbf{CI-Diff} and diffusion baselines on RareBench. Where L denotes LLM score and U denotes User Study. Our CI-Diff attains optimal LLM and user study results against all baseline methods.}
	\label{tab:LLM_human}
	%	\small
	%	\renewcommand{\arraystretch}{0.9} % 行高
	%	\setlength{\tabcolsep}{4pt}       % 列间距
	\resizebox{0.9\textwidth}{!}{
		\begin{tabular}{l|cc|cc|cc|cc|cc|cc|cc|cc|cc|cc}
			\toprule
			\textbf{Models} & \multicolumn{10}{c|}{\textbf{Single Object}} & \multicolumn{6}{c|}{\textbf{Multi Objects}} & \multicolumn{4}{c}{\textbf{Extend}} \\
			& \multicolumn{2}{c|}{Property} & \multicolumn{2}{c|}{Shape} & \multicolumn{2}{c|}{Texture} & \multicolumn{2}{c|}{Action} & \multicolumn{2}{c|}{Complex} & \multicolumn{2}{c|}{Concat} & \multicolumn{2}{c|}{Relation} & \multicolumn{2}{c|}{Complex} & \multicolumn{2}{c|}{Style} & \multicolumn{2}{c}{Scene} \\
			& L & U & L & U & L & U & L & U & L & U & L & U & L & U & L & U & L & U & L & U \\
			\midrule
			% --- Group 1: SDXL Pair ---
			SDXL & 66.5 & 72.5 & 69.5 & 80.0 & 75.0 & 73.5 & 56.5 & 63.5 & 79.0 & 71.0 & 59.5 & 72.0 & 46.0 & 69.5 & 68.0 & 69.0 & 93.0 & 79.0 & 64.0 & 68.0 \\
			\textbf{Our+SDXL} & \textbf{82.0} & \textbf{76.0} & \textbf{70.5} & \textbf{86.0} & \textbf{84.5} & \textbf{79.5} & \textbf{72.5} & \textbf{70.0} & \textbf{80.0} & \textbf{74.5} & \textbf{60.0} & \textbf{75.0} & \textbf{52.5} & \textbf{72.5} & \textbf{68.5} & \textbf{75.5} & \textbf{95.5} & \textbf{82.5} & \textbf{65.0} & \textbf{76.0} \\
			\midrule
			% --- Group 2: RealV Pair ---
			RealVisXL & 78.0 & 73.0 & 83.0 & 71.5 & 75.0 & 76.0 & 68.5 & 63.0 & 78.5 & 69.5 & 57.0 & 71.0 & 55.0 & 69.0 & 69.5 & 74.5 & 92.5 & 73.5 & 64.5 & 72.0 \\
			\textbf{Our+RealVisXL} & \textbf{85.0} & \textbf{81.0} & \textbf{85.0} & \textbf{87.0} & \textbf{81.0} & \textbf{84.5} & \textbf{75.5} & \textbf{70.5} & \textbf{85.0} & \textbf{80.5} & \textbf{58.0} & \textbf{76.5} & \textbf{59.5} & \textbf{75.0} & \textbf{73.0} & \textbf{78.0} & \textbf{94.0} & \textbf{80.0} & \textbf{71.0} & \textbf{76.0} \\
			\midrule
			% --- Group 3: SOTA Comparison ---
			SD1.5 & 58.5 & 45.5 & 58.5 & 46.5 & 49.0 & 48.5 & 44.0 & 37.5 & 52.0 & 49.0 & 37.5 & 36.0 & 35.5 & 40.5 & 33.5 & 36.5 & 80.5 & 67.5 & 46.0 & 65.5 \\
			Flux & 78.5 & 79.0 & 81.0 & 81.1 & 63.0 & 68.5 & 75.0 & 69.0 & 85.5 & 77.0 & 76.5 & 77.5 & 75.5 & 72.5 & 82.5 & 79.0 & 94.5 & 80.0 & 68.0 & 77.5 \\
			PixArt-$\alpha$ & 69.0 & 78.0 & 77.5 & 75.5 & 80.5 & 73.5 & 83.0 & 66.5 & 80.5 & 73.5 & 53.0 & 71.0 & 51.0 & 69.5 & 68.0 & 74.0 & 95.0 & 73.5 & 65.5 & 73.0 \\
			SynGen & 76.5 & 78.0 & 72.0 & 78.5 & 51.0 & 50.0 & 53.0 & 61.0 & 75.0 & 74.5 & 46.5 & 72.5 & 39.0 & 63.0 & 46.5 & 62.5 & 84.0 & 77.5 & 62.0 & 69.5 \\
			RPG & 62.0 & 67.5 & 59.5 & 50.0 & 73.5 & 55.0 & 51.0 & 62.0 & 74.0 & 58.0 & 52.0 & 51.5 & 44.0 & 42.5 & 61.5 & 40.5 & 91.0 & 69.5 & 64.5 & 63.0 \\
			SD3.0 & 70.0 & 77.5 & 87.5 & 73.0 & 62.0 & 78.0 & 71.0 & 67.0 & 80.0 & 77.0 & 68.0 & 79.5 & 72.0 & 73.5 & 78.0 & 75.0 & 92.5 & 77.5 & 61.5 & 72.0 \\
			R2F+SD3.0 & 78.0 & 79.0 & 79.5 & 75.0 & 76.5 & 82.0 & 70.5 & 72.5 & 84.0 & 81.0 & 68.5 & 84.5 & 63.0 & 77.0 & 80.5 & 78.0 & 92.5 & 80.0 & 61.0 & 75.0 \\
			SD3.5 & 80.0 & 81.0 & 78.5 & 71.0 & 77.0 & 73.5 & 76.5 & 72.0 & 84.0 & 81.0 & 77.0 & 88.0 & 74.0 & 79.5 & 86.0 & 79.0 & 91.5 & 80.5 & 64.5 & 74.0 \\
			\textbf{Our+SD3.5} & \textbf{\textcolor{red}{93.0}} & \textbf{\textcolor{red}{91.5}} & \textbf{\textcolor{red}{89.5}} & \textbf{\textcolor{red}{92.0}} & \textbf{\textcolor{red}{90.5}} & \textbf{\textcolor{red}{91.0}} & \textbf{\textcolor{red}{91.0}} & \textbf{\textcolor{red}{93.0}} & \textbf{\textcolor{red}{91.5}} & \textbf{\textcolor{red}{91.0}} & \textbf{\textcolor{red}{87.0}} & \textbf{\textcolor{red}{95.5}} & \textbf{\textcolor{red}{76.5}} & \textbf{\textcolor{red}{85.0}} & \textbf{\textcolor{red}{88.0}} & \textbf{\textcolor{red}{88.5}} & \textbf{\textcolor{red}{96.0}} & \textbf{\textcolor{red}{97.5}} & \textbf{\textcolor{red}{74.5}} & \textbf{\textcolor{red}{87.5}} \\
			\bottomrule
		\end{tabular}
	}
\end{table*}

\subsection{Temporal Morphological Fidelity Anchoring}
\label{sec:tmfa}

As discussed in Sec.~\ref{sec:concept_guidance}, the counterfactual guidance mechanism effectively alleviates the weakening of unusual attributes. However, during the actual denoising process, a large $s_{\text{rare}}$ is often required to thoroughly enhance unusual attributes. Such high-intensity feature injection to activate these attributes easily breaks the shape of the target object, leading to severe shape distortions. To resolve this contradiction---enhancing unusual attributes while effectively maintaining the shape consistency of the object---this paper proposes the \textbf{T}emporal \textbf{M}orphological \textbf{F}idelity \textbf{A}nchoring \textbf{(TMFA)} strategy.

The core idea of this strategy is to introduce a prior reference image to provide precise contour constraints for the generation process. To avoid background details and inherent texture interference caused by directly injecting raw RGB images \cite{anydoor}, this paper extracts an edge map from a background-removed reference image as the conditional input. Furthermore, to prevent contour constraints from interfering with early global semantic layout initialization and late micro-texture generation \cite{shixiong2}, we strictly implement the edge condition injection within a specific timestep interval $[\tau_1 T, \tau_2 T]$. The mathematical expression is:
\begin{equation}
	C_{\text{img}}(t) =
	\begin{cases} 
		\text{Edge}(\text{RemoveBG}(I_{\text{ref}})), & \text{if } t \in [\tau_1 T, \tau_2 T] \\
		0, & \text{otherwise}
	\end{cases},
	\label{eq:tmfa_condition}
\end{equation}
where $I_{\text{ref}}$ denotes the prior reference image. $\text{RemoveBG}(\cdot)$ utilizes a salient object detection (SOD) model \cite{SOD} to accurately separate the complex background, and $\text{Edge}(\cdot)$ applies the Canny algorithm \cite{edge} to extract the structural contours of the entity. 
Through this precise temporal intervention mechanism, the \textbf{TMFA} strategy effectively anchors the structure of the object during the core denoising stage, while unleashing the generation freedom at both the early and late timesteps. This ensures that unusual attributes achieve sufficient visual expression without causing shape distortions.

\section{Experiments}
\label{sec:experiments}

\subsection{Experimental Settings}
\label{subsec:settings}

\begin{figure*}[h]
	\centering
	\includegraphics[width=0.7\linewidth]{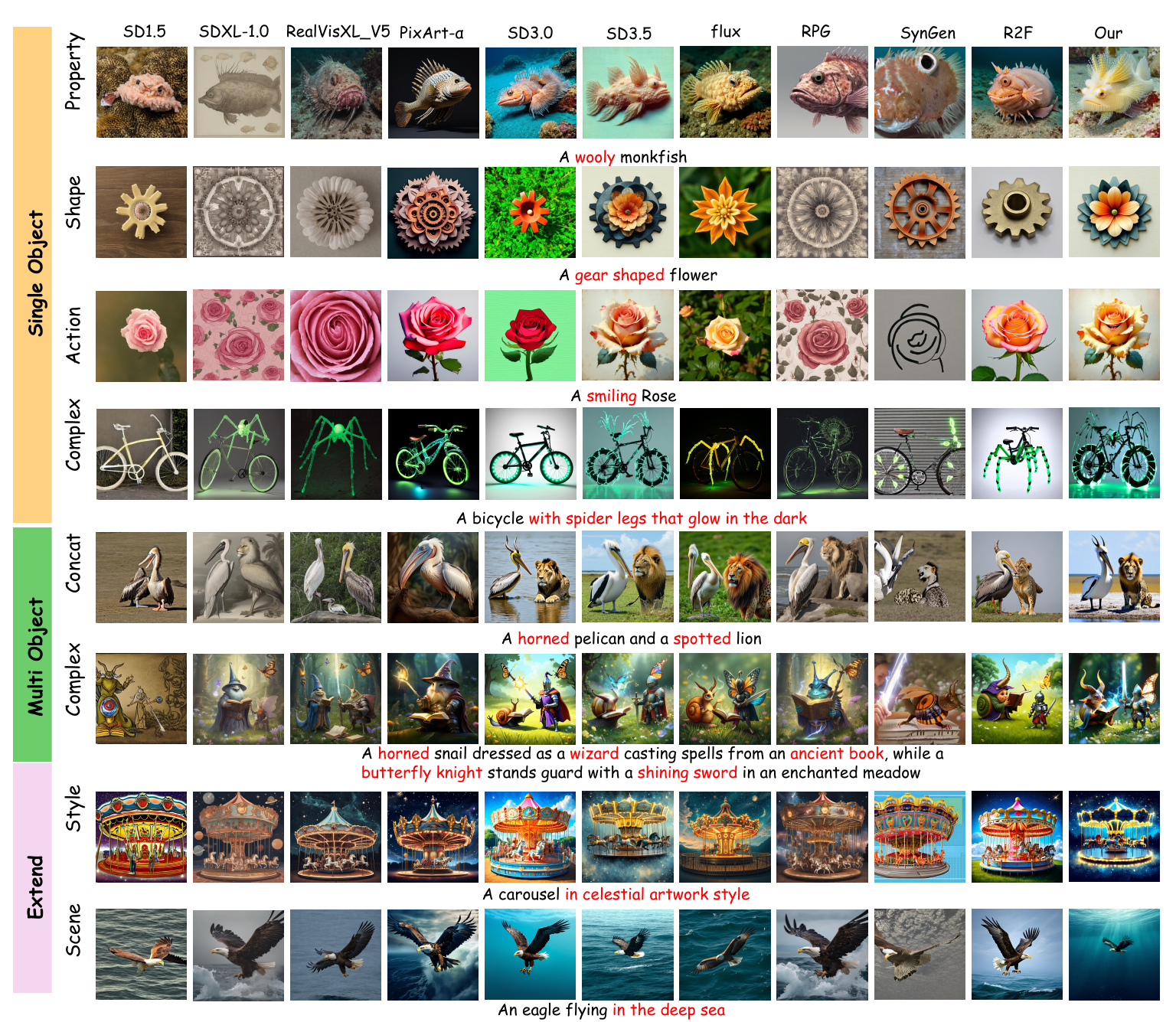}
	\caption{Qualitative comparison of \textbf{CI-Diff} with state-of-the-art diffusion baselines on RareBench. R2F can roughly generate unusual attributes yet leads to severe object deformation (e.g., distorted bicycles in the Complex category); pre-trained models like SD3.5 preserve complete object shapes but fail to fully express unusual attributes (e.g., the rose lacks smiling facial features in the Action category). By contrast, our CI-Diff successfully generates accurate rare attributes while maintaining intact object morphology across all categories.}
	\Description{Qualitative comparison showing our method outperforms baselines in structural integrity and rare attribute generation.}
	\label{fig:dingxingfenxi}
\end{figure*}

We evaluate \textbf{CI-Diff} on datasets covering both single-object and multi-object rare concept generation. We adopt RareBench \cite{R2F}, the latest benchmark for rare concept alignment, comprising five single-concept categories \textbf{(property, shape, texture, action, and complex)} and three multi-concept combinatorial categories \textbf{(concat, relation, and complex)}, with 40 text prompts per category. To further validate the generalization and superiority of our method across broader tasks, we additionally extend our evaluation to two test cases: \textbf{style} and \textbf{scene}. For evaluation metrics, we use \textbf{CLIP-T} \cite{clip} and \textbf{HPSv2} \cite{hpsv2} for text-image alignment, along with \textbf{LLM scores} and \textbf{User Study} to assess the generation quality of unusual attributes. All experiments are implemented in PyTorch on an NVIDIA A40 GPU. \textbf{CI-Diff} is plug-and-play and compatible with all mainstream diffusion models; we use SD3.5 \cite{SD3035} by default, 30 steps, and official hyper-parameters for fair comparison. Unless otherwise specified, the rare guidance scale is set to $s_{\text{cfg}} = 5$ and the image intervention window for \textbf{TMFA} is set to $[\tau_1, \tau_2] = [0.1, 0.9]$. \textbf{\textit{(Due to page limitations, more discussions about the implementation details are provided in Sec. ~\ref{Appendix} of the Appendix.)}}

\subsection{Comparison with State-of-the-Art Methods}
\label{subsec:Comparison}
\noindent \textbf{Quantitative Comparison.} To validate the superiority of \textbf{CI-Diff}, we perform a thorough comparison against state-of-the-art diffusion models for rare concept generation. We consider two groups of representative methods. The first group includes SD1.5 \cite{LDM_sd15}, SDXL-1.0 \cite{sdxl} (along with its fine-tuned version RealVisXL\_V5), PixArt-$\alpha$ \cite{pix}, FLUX-schnell, and SD3.0 \cite{SD3035} (along with its fine-tuned version SD3.5). These models are pre-trained on massive conventional datasets, which leads to a strong \textbf{common knowledge bias} within the models. The second group consists of SynGen \cite{syngen}, RPG \cite{RPG}, and R2F \cite{R2F}, which primarily enhance unusual attributes of objects  by controlling external guidance, but often introduce other problems during the generation process.The quantitative results summarized in Table.~\ref{tab:clip_hpsv2} and Table.~\ref{tab:LLM_human} highlight our findings: \textbf{CI-Diff} outperforms its competitors across CLIP-T, HPSv2, LLM scores, and User Study metrics, achieving superior performance. Notably, the high LLM-based evaluation scores directly verify the outstanding advantage of our method in decoupling and enhancing the unusual attributes of objects. It is noteworthy that when \textbf{CI-Diff} is combined with SDXL-1.0 (along with its fine-tuned version RealVisXL\_V5) or SD3.5, all metrics show significant improvements. This demonstrates that the core contribution of \textbf{CI-Diff} lies in unleashing and activating the model's potential to generate unusual attributes, rather than solely relying on the parameter scale of the underlying base models. Furthermore, in two extended general task categories, namely Style and Scene, the proposed method consistently surpasses all comparative methods. This sufficiently validates the strong generalization capability of \textbf{CI-Diff}, enabling it to effectively adapt to many downstream tasks in the broader text-to-image generation domain.

\begin{figure*}[h]
	\centering
	\includegraphics[width=0.8\linewidth]{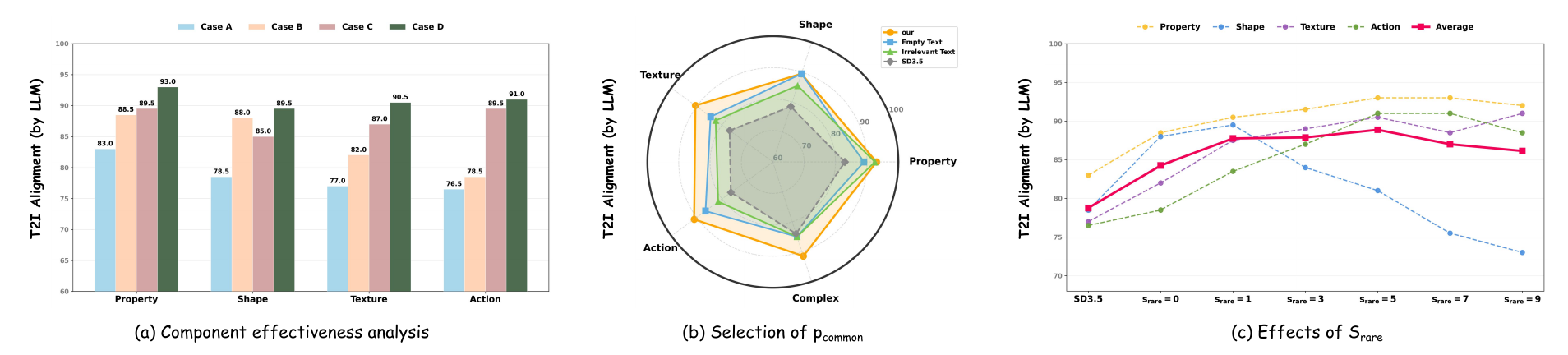}
	\caption{Ablation studies on RareBench. We conduct comprehensive ablation experiments on the RareBench benchmark to verify the effectiveness of each core module of our CI-Diff, with LLM score adopted as the evaluation metric.}
	\Description{Three subfigures showing ablation studies.}
	\label{fig:santu}
\end{figure*}

\noindent \textbf{Qualitative Comparison.} To shed further more light on the observations, Fig.~\ref{fig:dingxingfenxi} showcases the visualization results of all compared methods on the RareBench dataset. It is evident that our proposed \textbf{CI-Diff} achieves superior performance in both the expression of unusual attributes and the shape consistency of the objects with the text prompts. Analysis reveals that images generated by foundation models pre-trained on massive conventional datasets, such as SDXL and FLUX-schnell, generally struggle to effectively manifest the unusual attributes of objects. This result validates our intuition in Sec.~\ref{sec:intro}: models pre-trained on massive conventional datasets are deeply influenced by a strong \textbf{common knowledge bias}, which suppresses the expression of unusual attributes. Although R2F \cite{R2F} attempts to guide the generation using common concepts, it also introduces a substantial amount of redundant information unrelated to unusual attributes, leading to shape distortions in the generated objects (e.g., the deformed monkfish in the property category and the distorted bicycle in complex). In contrast, \textbf{CI-Diff} successfully and naturally activates the synthesis of unusual attributes (e.g., a "smiling rose" in the action category) while perfectly maintaining the shape consistency of the generated objects. \textbf{\textit{(Due to page limitations, see more high-resolution compared results in Fig.~\ref{fig:dingxingfenxi2} and Sec. ~\ref{Results} of the Appendix .)}}

\subsection{Ablation Study}
\label{sec:ablation}

\noindent \textbf{Discussion on Different Modules of \textbf{CI-Diff}.} To validate the effectiveness of various modules in our \textbf{CI-Diff}, we perform an ablation study on the RareBench dataset with several variants: \textbf{Case A:} the pre-trained base model SD3.5; \textbf{Case B:} removing Eq.~(\ref{eq:final_guidance}) and adopting Eq.~(\ref{eq:cfg_guidance}) to calculate the predicted noise $\epsilon_\theta$; \textbf{Case C:} removing the \textbf{TMFA} module from \textbf{CI-Diff}; \textbf{Case D:} the full \textbf{CI-Diff} method. As illustrated in Fig.~\ref{fig:santu}(a), our full \textbf{CI-Diff} model significantly outperforms Case A in terms of the LLM score, confirming that our \textbf{CI-Diff} can successfully decouple unusual attributes from text prompts and enhance them without compromising the original shape of the object.

\noindent \textbf{Selection of $\boldsymbol{p_{\text{common}}}$.} This paper designs three types of common text prompts to compare with rare text prompts, aiming to explore which common text prompt can best achieve the decoupling of unusual attributes: 1) empty text; 2) irrelevant text (e.g., when the rare text prompt is ``a hairy frog'', the common text prompt is set to ``a car''); 3) subject-aligned common text (e.g., ``a frog''). As shown in Fig.~\ref{fig:santu}(b), all three types of common text prompts improve the performance compared with the baseline model SD3.5. Nevertheless, the third type of text prompt achieves the best overall LLM scores. This demonstrates that by contrasting rare concepts with their corresponding subject-level common concepts, the model can successfully decouple unusual attributes, which further validates the effectiveness of Eq.~(\ref{eq:nde_noise_pure}).
\begin{figure}[h]
	\centering
	\includegraphics[trim=0 0 0 0bp, clip, width=0.9\linewidth]{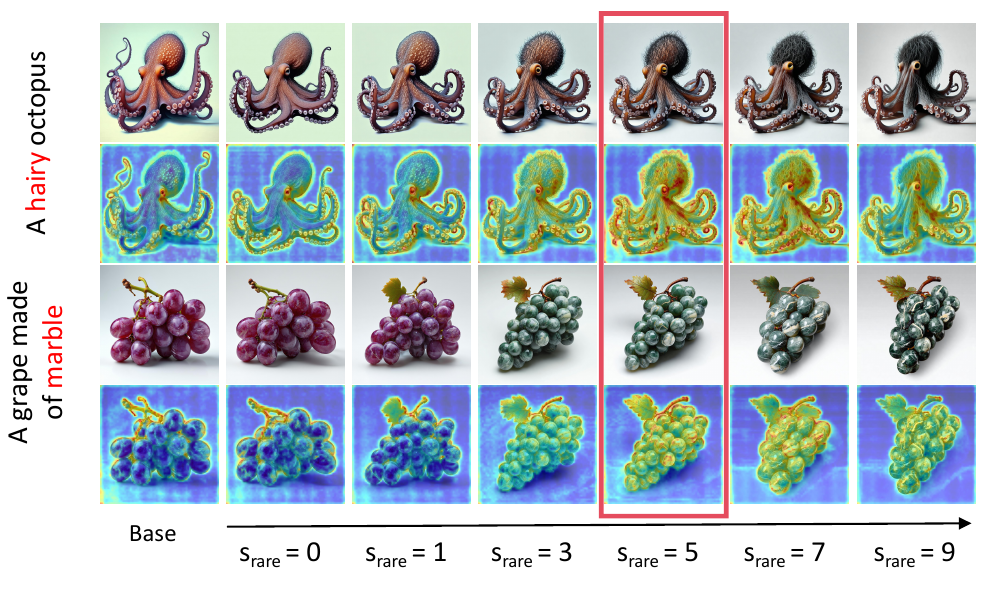}
	\caption{Visualizing the impact of $s_{rare}$ on generated images and cross-attention maps. Different values of $s_{rare}$ greatly change the expression of rare attributes and the focus range of cross-attention, and visual comparison proves that $s_{rare}$ = 5 achieves the optimal generation result.}
	\Description{Attention maps visualization for varying guidance scale values.}
	\label{fig:srarezhuyili}
\end{figure}

\noindent \textbf{Effects of $s_{\text{rare}}$.} To evaluate the impact of the rare guidance coefficient $s_{\text{rare}}$, we vary its value from 0 to 9 and measured the LLM scores for each setting. As illustrated in Fig.~\ref{fig:santu}(c), the model achieved the optimal average performance across all categories at $s_{\text{rare}} = 5$. To explore the underlying reasons for this phenomenon, we visualized the attention maps in Fig.~\ref{fig:srarezhuyili} for analysis. The results show that at $s_{\text{rare}} = 5$, the model can steadily and accurately focus on the correct regions corresponding to unusual attributes such as ``hairy'' and ``marble''. Nevertheless, an excessively large $s_{\text{rare}}$ causes the over-expression of unusual attributes, thereby suppressing the shape of objects. For instance, the morphological of the octopus becomes extremely blurry when $s_{\text{rare}} = 9$.  In addition, for text prompts of the ``Shape'' category, $s_{\text{rare}} = 1$ achieves the best performance. This is attributed to the fact that the Temporal Morphological Fidelity Anchoring strategy \textbf{(TMFA)} can effectively constrain and maintain the shape consistency of objects. \textbf{\textit{(Due to page limitations, more experimental analyses can be found in Sec. ~\ref{Additional} of the Appendix.)}}

\section{Conclusion}
\label{sec:Conclusion}
In this paper, we target the \textbf{common knowledge bias} issue in rare concept generation for diffusion models. Technically, we propose the counterfactual inference-based diffusion approach \textbf{CI-Diff}, which first introduces causal inference into the text-to-image rare concept generation task. By constructing causal graphs and counterfactual scenarios, we extract the natural direct effect to decouple unusual attributes from rare concepts, and reformulate the classifier-free guidance mechanism to map the causal effect of unusual attributes into the noise space for enhancement. Meanwhile, the Temporal Morphological Fidelity Anchoring strategy is devised to inject edge priors at specific timesteps, ensuring the shape consistency of generated objects while strengthening unusual attribute expression. Extensive experiments on the RareBench benchmark demonstrate the superiority of our training-free plug-and-play \textbf{CI-Diff} over state-of-the-art methods.

\noindent \textbf{Acknowledgments} This research is supported by Institute of Advanced Medicine and Frontier Technology (2023IHM01080), and sponsored by CCF-NetEase ThunderFire Innovation Research Funding (NO. CCF-Netease 202513); The computation is completed on the HPC Platform of Hefei University of Technology.

%\begin{acks}
%To Robert, for the bagels and explaining CMYK and color spaces.
%\end{acks}

\clearpage

\bibliographystyle{ACM-Reference-Format}
\bibliography{sample-base}

\newpage
\appendix
\onecolumn

\section{Appendix}
\label{Appendix}
Due to page limitation of the mainbody, as indicated by our submission, the appendix offers further technical analysis, implementation details, and more qualitative results, which are summarized below:
{
	\leftmargini=5mm
	\begin{itemize}
		\item \textbf{Detailed implementation and experimental settings}. including the CI-Diff algorithm (Algorithm~\ref{alg:cidiff}), the construction of extended evaluation categories, LLM-based scoring criteria, and the user study setup, as mentioned in Sec.~\ref{subsec:settings} of the mainbody.  (Sec.~\ref{Appendix}).
		\item \textbf{Additional qualitative analysis for rare concept generation}, including more generation results of CI-Diff on the RareBench dataset and further comparisons with state-of-the-art (SOTA) methods, as mentioned in Sec.~\ref{subsec:Comparison} of the mainbody . (Sec. ~\ref{Results}).
		\item \textbf{Additional Ablation Study}, including the exploration of the injection timestep interval $[\tau_1 T, \tau_2 T]$ in the TMFA strategy and the quantitative evaluation of inference efficiency (latency and memory usage), as mentioned in Sec.~\ref{sec:ablation} of the mainbody. (Sec.~\ref{Additional}).
		\item \textbf{Additional Discussions}, including the clarification of the construction rule for $p_{common}$ and the analysis of inherent limitations of our proposed method. (Sec.~\ref{re}).
	\end{itemize}
}

\begin{algorithm*}[h]
	\caption{\textbf{CI-Diff}}
	\label{alg:cidiff}
	\begin{algorithmic}[1]
		\renewcommand{\algorithmicrequire}{\textbf{Input:}}
		\renewcommand{\algorithmicensure}{\textbf{Output:}}
		\REQUIRE $p_{\text{rare}}$: rare text prompt; $p_{\text{common}}$: common text prompt; $I_{\text{ref}}$: subject reference image; $s_{\text{cfg}}$: CFG guidance scale; $s_{\text{rare}}$: rare guidance scale; $S$: random seed; $DM$: Stable Diffusion model; $\tau_1, \tau_2$: injection interval.
		\ENSURE $I_{\text{gen}}$: generated image aligned with the rare text prompt.
		\medskip
		
		\STATE Sample standard Gaussian noise $z_T \sim \mathcal{N}(0, 1)$ using random seed $S$;
		\STATE $I_{\text{ref}}^* \leftarrow \text{Edge}(\text{RemoveBG}(I_{\text{ref}}))$; \hfill // Extract morphological structure
		
		\FOR{$t = T, T-1, \dots, 1$}
		\IF{$t \in [\tau_1 T, \tau_2 T]$}
		\STATE $C_i \leftarrow I_{\text{ref}}^*$; \hfill // Inject edge constraints
		\ELSE
		\STATE $C_i \leftarrow \text{None}$;
		\ENDIF
		
		\STATE $\epsilon_{\text{rare}}, \epsilon_{\text{normal}}, \epsilon_{\emptyset} \leftarrow$ Predict noises via $DM(z_t, t, p_{\text{rare}}, C_i)$ and $DM(z_t, t, p_{\text{common}}, C_i)$;
		
		\STATE $\hat{\epsilon}_{\text{CFG}} \leftarrow s_{\text{cfg}} \cdot \epsilon_{\text{rare}} + (1 - s_{\text{cfg}}) \cdot \epsilon_{\emptyset}$; \hfill // Standard CFG
		\STATE $\hat{\epsilon}_{\text{NDE}} \leftarrow s_{\text{rare}} \cdot (\epsilon_{\text{rare}} - \epsilon_{\text{normal}})$; \hfill // Counterfactual Accentuation
		\STATE $\tilde{\epsilon}_\theta \leftarrow \hat{\epsilon}_{\text{CFG}} + \hat{\epsilon}_{\text{NDE}}$;
		\STATE $z_{t-1} \leftarrow \text{SamplerStep}(z_t, \tilde{\epsilon}_\theta, t)$; \hfill // Update latent
		\ENDFOR
		
		\STATE \textbf{Return} $I_{\text{gen}} \leftarrow \text{Decoder}(z_0)$;
	\end{algorithmic}
\end{algorithm*}

\noindent \textbf{Construction of Extended Evaluation Categories.} To verify the generalization capability of \textbf{CI-Diff}, we leverage the multimodal large language model \textbf{Qwen2.5-VL-7B-Instruct} to construct two novel categories, \textbf{Style} and \textbf{Scene}, by mimicking the rare concept of the RareBench \cite{R2F} benchmark. Each category consists of 40 text prompts, maintaining a sample scale consistent with the original RareBench categories. The specific instructions provided to the model are as follows:

\textit{You are my assistant responsible for imitating the style of the RareBench rare concept generation dataset to construct 40 high-quality and diverse text prompts for text-to-image generation tasks, focusing on rare/uncommon styles and scenes. Please provide specific text descriptions according to the following requirements:}
{
	\leftmargini=5mm
	\begin{itemize}
		\item \textit{Style text prompts: Each text prompt focuses on a rare, niche, and distinctive artistic, painting, rendering, or visual expression style, avoiding common popular styles. Examples: A playful fox cub in pixel art game style; An elephant in wooden sculpture.}
		\item \textit{Scene text prompts: Each text prompt describes a rare, unconventional, and challenging scene with details of space, atmosphere, and object relationships, avoiding ordinary daily scenes. Examples: A penguin waddling in the desert; A bee building a nest in a space capsule.}
	\end{itemize}
}
\begin{figure*}[h]
	\centering
	\includegraphics[trim=0 0 0 0bp, clip, width=0.8\linewidth]{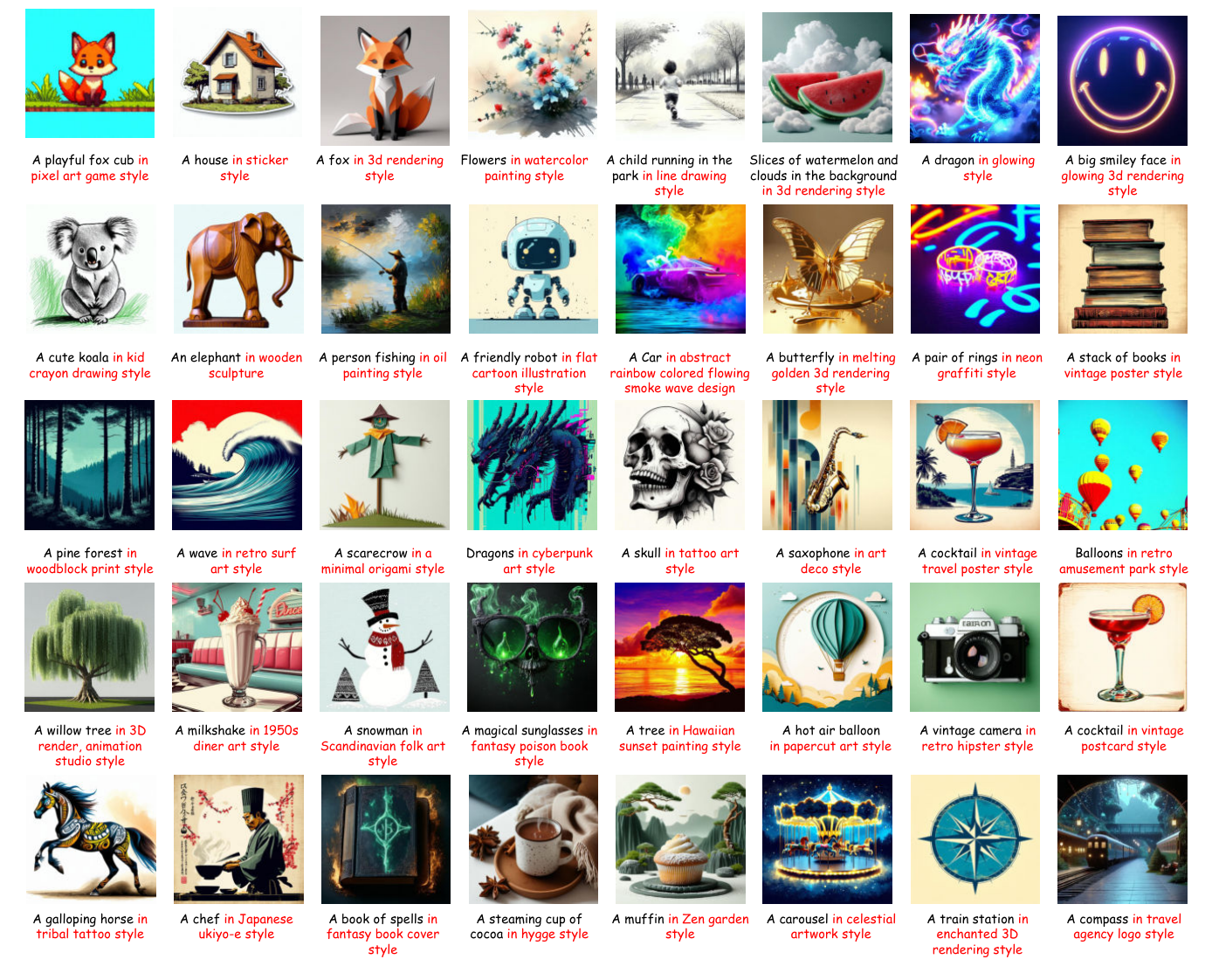}
	\caption{Qualitative results of CI-Diff on the extended \textbf{Style} category. Our method effectively captures rare and unconventional artistic styles while maintaining subject integrity.}
	\Description{Visual generation results demonstrating rare style concepts.}
	\label{fig:fengge}
\end{figure*}
\noindent The qualitative generation results of \textbf{CI-Diff} on these two extended categories are illustrated in Fig.~\ref{fig:fengge} and Fig.~\ref{fig:changjin}. As depicted in Fig.~\ref{fig:fengge}, \textbf{CI-Diff} excels in rendering diverse niche artistic styles with high precision, such as a house in sticker style or A scarecrow in a minimal origami style.  For the \textbf{Scene} category, as shown in Fig.~\ref{fig:changjin}, our method successfully synthesizes high-quality images even when the subjects and environments are highly unconventional or conflicting (e.g., A penguin waddling in
the desert or A mole digging holes in the high-altitude clouds). These results further substantiate the arguments presented in Sec.~\ref{subsec:Comparison}, demonstrating that \textbf{CI-Diff} is not only effective for rare concept generation but also is versatile for various downstream tasks within the broader text-to-image generation landscape.

\noindent \textbf{LLM Scoring.} Many existing evaluation metrics fail to accurately assess the specific unusual attributes embedded within rare concepts. For instance, standard metrics such as CLIP-T and HPSv2 primarily focus on overall text-image alignment but lack the capability to evaluate unusual attributes explicitly. To address this, we leverage the advanced multimodal model \textbf{Qwen2.5-VL-7B-Instruct} as an automated evaluator, focusing on two critical dimensions: the presentation degree of unusual attributes and the structural integrity of the main subject. Given the generated image and the corresponding rare text prompt, the model is required to assign a score from 1 to 5 according to the criteria defined below. These scores are ultimately linearly mapped to a 0--100 scale. The evaluation prompt is provided as follows:

\textit{You are my assistant to evaluate the correspondence of an image to a given text prompt. Focus on the objects in the image and their attributes (such as color, shape, texture), spatial layout, and action relationships. Evaluate how well the image aligns with the text prompt based on the following scale:}
{\leftmargini=5mm
	\begin{itemize}
		\item \textit{5: The image perfectly matches the content of the text prompt, with no discrepancies.}
		\item \textit{4: The image portrays most of the actions, events, and relationships, but with minor discrepancies.}
		\item \textit{3: The image depicts some elements from the text prompt, but ignores key parts or details.}
		\item \textit{2: The image does not depict any actions or events that match the text.}
		\item \textit{1: The image fails entirely to convey the scope of the text prompt.}
	\end{itemize}
}

\noindent \textbf{User Study.} To further evaluate the representation of unusual attributes, we conducted a comprehensive user study. We randomly selected 5 texts from each of the eight original categories in RareBench (Property, Shape, Texture, Action, Single-object Complex, Concat, Relation, and Multi-object Complex) and our two extended categories (Style and Scene), resulting in a total of 50 evaluation samples. Ten independent volunteers were invited to participate in the study. Each participant was tasked with scoring the generated images on a scale of 1 to 5 based on two key criteria: 1) Does the image accurately present the unusual attributes described in the prompt? 2) Is the main subject's structure clear and free from unreasonable morphological distortions? Finally, to ensure a standardized comparison, these 1--5 ratings were linearly mapped to a 0--100 scale.

\section{Additional qualitative analysis for rare concept generation}
\label{Results}

\noindent \textbf{Extended Qualitative Analysis on the SD3.5 Backbone.} We provide additional qualitative generation results of the proposed \textbf{CI-Diff} method using the SD3.5 backbone in Fig.~\ref{fig:diff_model}. These visualizations, spanning various categories in the RareBench dataset, further underscore the robust performance and superior expressive capability of our method in accurately manifesting unusual attributes within rare concept generation tasks.
\begin{figure*}[h]
	\centering
	\includegraphics[trim=0 0 0 0bp, clip, width=0.8\linewidth]{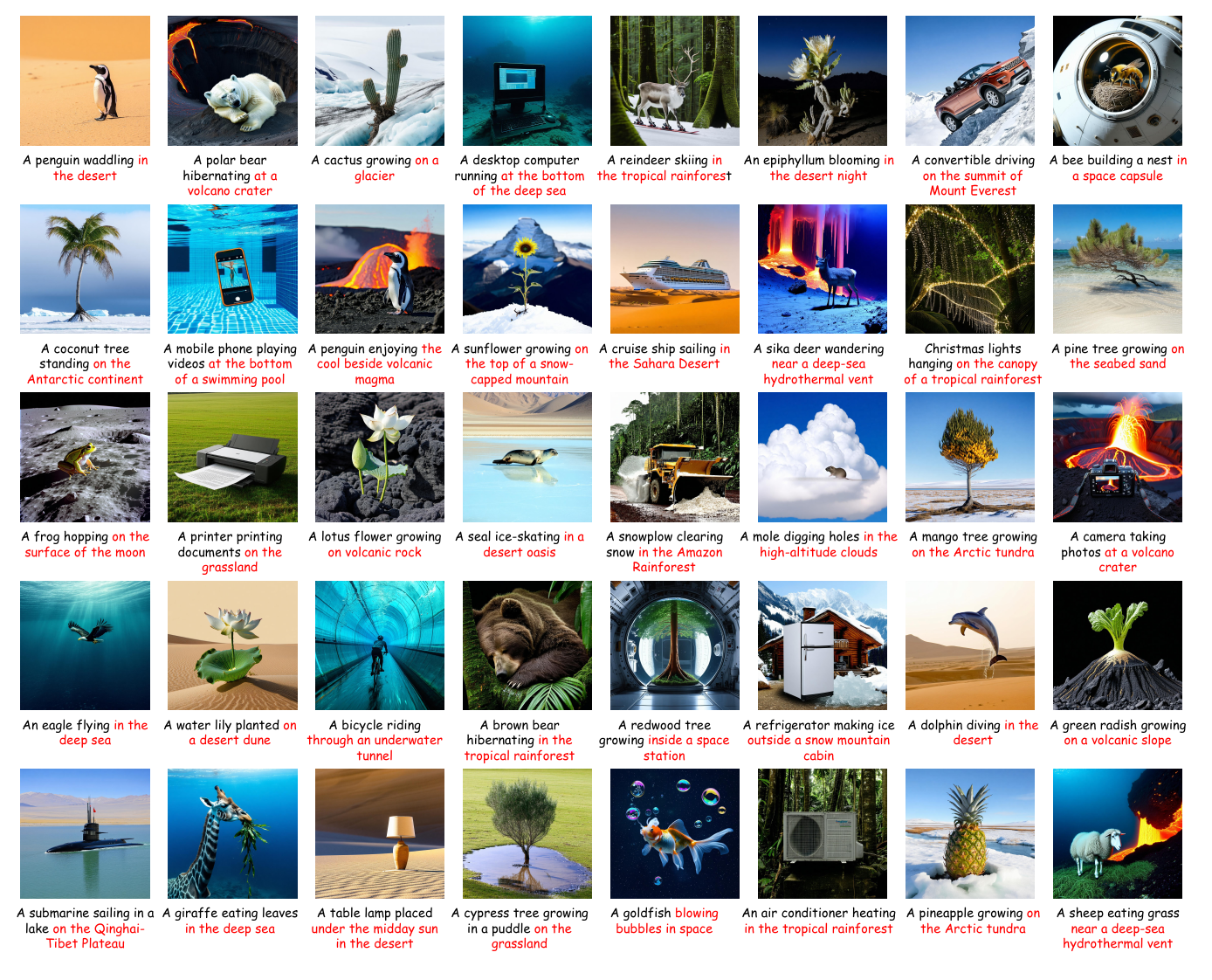}
	\caption{Qualitative results of \textbf{CI-Diff} on the extended \textbf{Scene} category. Our method successfully generates complex and unusual environmental scenes while achieving superior text-image alignment even in complex and unusual scenes.}
	\Description{Visual generation results demonstrating rare scene concepts.}
	\label{fig:changjin}
\end{figure*}

\noindent \textbf{Qualitative Comparison on the SDXL Backbone.} For a fair comparative analysis, we unify the backbone model to SDXL for all competing methods. The qualitative results of vanilla SDXL \cite{sdxl}, RPG-SDXL \cite{R2F}, R2F-SDXL \cite{RPG}, and our \textbf{CI-Diff} (based on SDXL) are illustrated in Fig.~\ref{fig:sdxl}. As shown in the figure, \textbf{CI-Diff} achieves superior generation performance among all compared approaches. Specifically, our method accurately renders unusual attributes; for instance, in the case of ``a hairless sheep,'' only \textbf{CI-Diff} successfully synthesizes the image as described. Furthermore, our approach effectively preserves the structural integrity and morphological consistency of the subjects. In contrast, for prompts such as ``a wooly banana'' and ``a zebra-striped duck,'' the images generated by SDXL, RPG, and R2F suffer from significant shape distortions, failing to maintain the essential structure of the main subjects.
\begin{figure*}[h]
	\centering
	\includegraphics[trim=0 0 0 0bp, clip, width=0.8\linewidth]{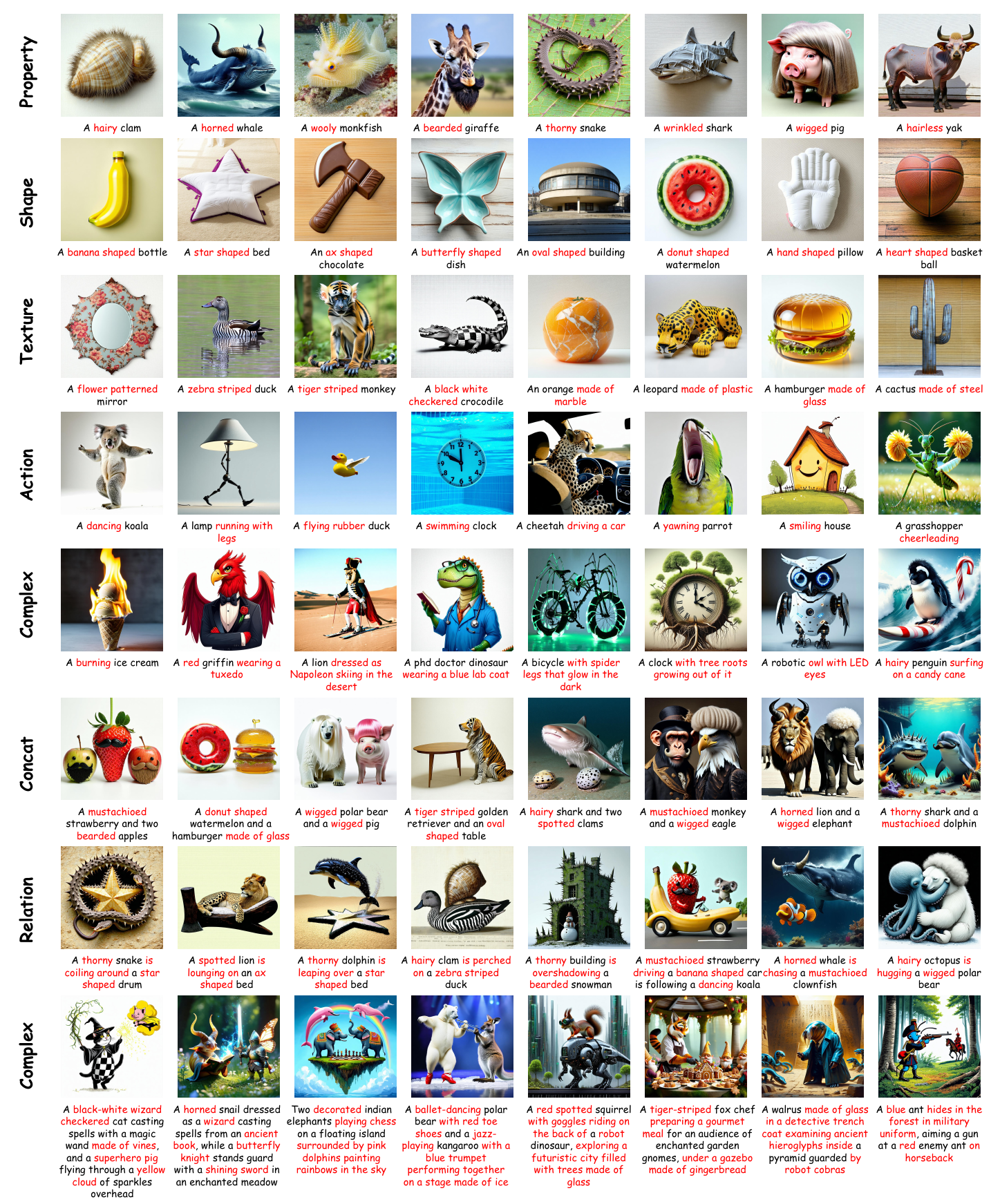}
	\caption{More comprehensive qualitative results of our CI-Diff method across various rare concept categories.}
	\Description{Additional generation results of CI-Diff on SD3.5.}
	\label{fig:diff_model}
\end{figure*}
\begin{figure*}[h]
	\centering
	\includegraphics[trim=0 0 0 0bp, clip, width=0.7\linewidth]{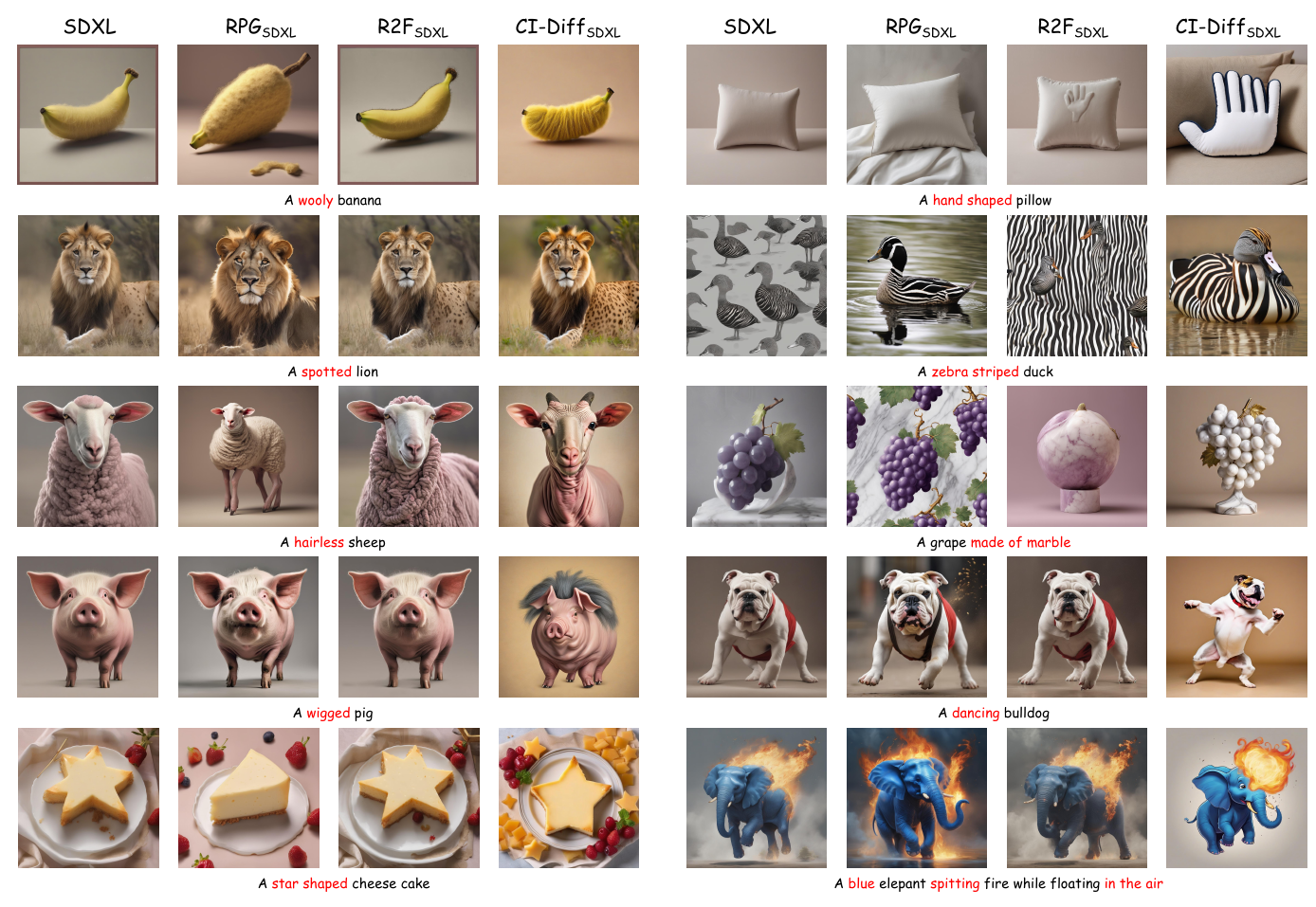}
	\caption{Qualitative comparison among vanilla SDXL, RPG-SDXL, R2F-SDXL, and our CI-Diff (SDXL backbone) on rare concept prompts.}
	\Description{Visual comparison of baseline methods and CI-Diff on the SDXL backbone.}
	\label{fig:sdxl}
\end{figure*}
\begin{figure*}[h]
	\centering
	\includegraphics[trim=0 0 0 0bp, clip, width=0.5\linewidth]{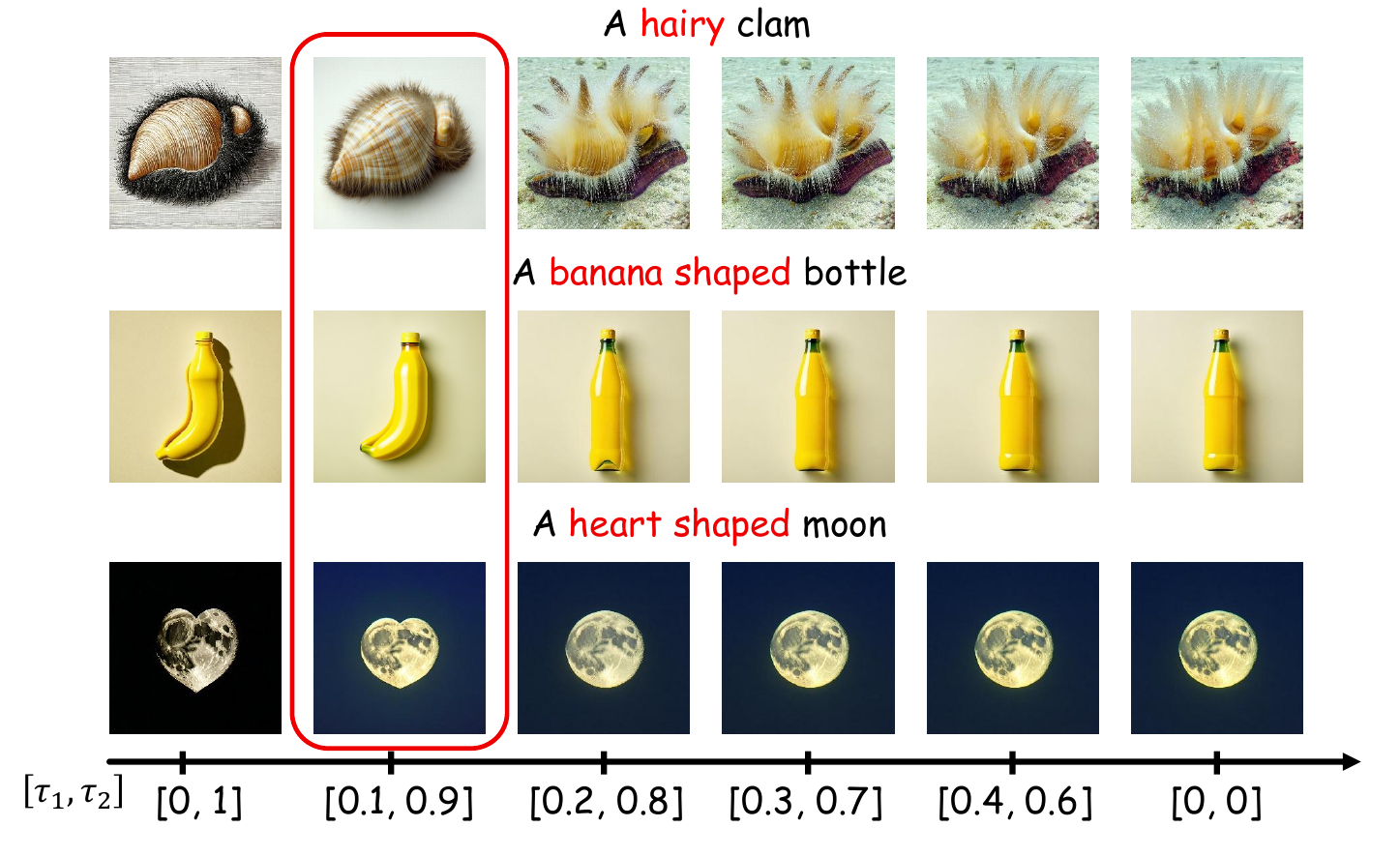}
	\caption{Qualitative ablation study on the TMFA injection intervals $[\tau_1, \tau_2]$.}
	\Description{Visual comparison of baseline methods and CI-Diff on the SDXL backbone.}
	\label{fig:tmfa}
\end{figure*}

\section{Additional Ablation Study}
\label{Additional}
\noindent \textbf{Inference Efficiency Analysis.} To further validate the efficiency advantages of our \textbf{CI-Diff} method, we conduct an additional ablation study focusing on inference performance. All methods in this evaluation are built upon the SDXL backbone to ensure fairness. We assess two critical metrics for practical deployment: average inference time per image and peak video memory (VRAM) consumption. As presented in Table.~\ref{tab:ablation_efficiency}, our \textbf{CI-Diff} achieves an excellent trade-off between generation quality and computational efficiency. Compared to the resource-heavy RPG+SDXL and R2F+SDXL, our method drastically reduces both time cost and memory usage. Compared to the vanilla SDXL, although \textbf{CI-Diff} introduces additional computational overhead, it successfully enables the generation of rare concepts, thereby proving its high practicality for real-world applications.

\begin{table}[h]
	\setlength{\abovecaptionskip}{0pt}
	\setlength{\belowcaptionskip}{0pt}
	\centering
	\caption{Ablation study on the inference efficiency of different methods.}
	\label{tab:ablation_efficiency}
	\small
	\renewcommand{\arraystretch}{0.8} % 统一和示例一样的行高
	\resizebox{0.8\linewidth}{!}{ % 宽度自适应行宽
		\begin{tabular}{lcc|lcc}
			\toprule
			\textbf{Method} & Time (s) & VRAM (GB) & \textbf{Method} & Time (s) & VRAM (GB) \\
			\midrule
			SDXL           & 6.38  & 10.49 & RPG + SDXL     & 32.62 & 35.14 \\
			R2F + SDXL     & 38.03 & 44.77  & CI-Diff + SDXL & \color{red}\textbf{12.94} & \color{red}\textbf{19.69} \\
			\bottomrule
		\end{tabular}
	}
\end{table}

\begin{table}[h]
	\setlength{\abovecaptionskip}{0pt}
	\setlength{\belowcaptionskip}{0pt}
	\centering
	\caption{Ablation study on alternative guidance strategies}
	\label{tab:table1}
	\small
	\renewcommand{\arraystretch}{0.8} % 行高
	\resizebox{\linewidth}{!}{ % 宽度根据你的需要调整
		\begin{tabular}{lccc|lccc}
			\toprule
			\textbf{Metric: LLM-Score $\uparrow$} & Property & Concat & Scen &\textbf{Metric: LLM-Score $\uparrow$} & Property & Concat & Scen \\
			\midrule
			SD 3.5 (Base)               & 80.0 & 77.0 & 64.5 &High CFG Scale ($s_{cfg}$=10)       & 84.5 & 74.5 & 63.5 \\
			Negative Prompt ($P_{common}$)            & 79.5 & 66.5 & 64.0 &Attention Boosting          & 85.5 & 82.5 & 69.5 \\
			\textbf{CI-Diff }                    & \color{red}\textbf{93.0} & \color{red}\textbf{87.0} & \color{red}\textbf{74.5} &- &- &- &- \\
			\bottomrule
		\end{tabular}
	}
\end{table}

\begin{table}[h]
	\centering
	\caption{Ablation study on image guidance types in \textbf{TMFA}.}
	\label{tab:TMFA_img}
	\resizebox{0.6\linewidth}{!}{
		\begin{tabular}{lccccc}
			\toprule
			Method & Img & Seg & Seg+Gray & Seg+Noise & Seg+Edge \\
			\midrule
			\textit{property} & 91.0 & 91.0 & 93.0 & 92.5 & 93.0 \\
			\textit{shape}    & 80.5 & 83.0 & 80.5 & 83.0 & 89.5 \\
			\textit{texture}  & 88.5 & 89.0 & 89.0 & 89.5 & 90.5 \\
			\textit{action}   & 89.0 & 90.5 & 89.0 & 89.5 & 91.0 \\
			\bottomrule
		\end{tabular}
	}
\end{table}

\begin{table}[h]
	\centering
	\caption{Ablation study on TMFA injection intervals $[\tau_1, \tau_2]$ across diverse rare concept categories. The scores are evaluated by LLM (0--100 scale).}
	\label{tab:ablation_tmfa}
	\resizebox{0.6\linewidth}{!}{
		\begin{tabular}{lcccccc}
			\toprule
			Interval & (0, 0) & (0, 1) & \textbf{(0.1, 0.9)} & (0.2, 0.8) & (0.3, 0.7) & (0.4, 0.6) \\
			\midrule
			\textit{property} & 89.5 & 91.0 & \textbf{93.9} & 90.5 & 89.5 & 89 \\
			\textit{shape}    & 85.0 & 86.5 & \textbf{89.5} & 76.5 & 72.5 & 72.5 \\
			\textit{texture}  & 87.0 & 90.0 & \textbf{90.5} & 86.5 & 86.5 & 87.5 \\
			\textit{action}   & 89.5 & 90.5 & \textbf{91.0} & 90.0 & 90.5 & 90.0 \\
			\bottomrule
		\end{tabular}
	}
\end{table}

\textbf{Different Guidance Strategies.} To verify the superiority of our counterfactual guidance scheme, we compare five distinct generation guidance strategies: (1) SD3.5 baseline; (2) taking the subject-aligned common prompt $p_{common}$ as negative prompt; (3) increasing the classifier-free guidance (CFG) scale; (4) amplifying the cross-attention weights corresponding to atypical attributes within rare prompts; (5) our full \textbf{CI-Diff} method. Table.~\ref{tab:table1} presents the quantitative comparison results among the three alternative guidance strategies and \textbf{CI-Diff}. As can be observed from the table, simply utilizing negative prompts to distinguish common and rare concepts or merely raising the CFG guidance scale cannot effectively decouple and enhance unusual attributes, and their generation performance is obviously inferior to \textbf{CI-Diff}. These results sufficiently demonstrate that our counterfactual-inference-based guidance strategy achieves superior performance on rare concept generation tasks.

\noindent  Importance of \textbf{TMFA}. This section mainly investigates the impact of injecting different types of image features in the \textbf{TMFA} module on the final generated images. We compare the following injection forms: original RGB images, background-removed RGB images, background-removed grayscale images, noise maps, and edge maps. The results of the ``Shape'' category in Table.~\ref{tab:TMFA_img} demonstrate that injecting background-removed edge maps achieves significantly better performance than other types of images. This confirms that during the image generation process, edge maps can not only effectively constrain the object shape but also do not interfere with the expression of unusual attributes, which is fully consistent with our theoretical analysis in Sec.~\ref{sec:tmfa}. 

\noindent \textbf{Ablation Study on TMFA Time-step Intervals.} We conduct a comprehensive qualitative and quantitative assessment of the injection time-step interval $[\tau_1 T, \tau_2 T]$ within the TMFA strategy. Six distinct interval configurations are investigated: $[\tau_1, \tau_2] \in$ \{ [0,1], [0.1, 0.9], [0.2, 0.8], [0.3, 0.7], [0.4, 0.6], [0,0]\}. As indicated by the quantitative results in Table~\ref{tab:ablation_tmfa}, the LLM-based composite score peaks within the $[0.1, 0.9]$ interval. Visualizations in Fig.~\ref{fig:tmfa} further demonstrate that imposing edge constraints throughout the entire denoising process (i.e., $[0, 1]$) compromises the overall generation quality. Conversely, excessively narrowing the interval (e.g., from $[0.2, 0.8]$ to $[0.4, 0.6]$) results in an insufficient effective duration for morphological anchoring, thereby failing to maintain structural stability.

\section{Additional Discussions}\label{re}
\noindent \textbf{The clarification on the construction of $P_{\text{common}}$.} As stated in Sec.~\ref{sec:counterfactual_reasoning}, \textit{rare text prompts generally consist of uncommon combinations of adjective attributes and noun subjects}. Accordingly, we can extract the noun subjects to construct the common prompt $P_{\text{common}}$. For instance, given the prompt ``\textit{A hairy frog is sitting on top of a spotted lizard}'',  $P_{\text{common}}$ is  ``\textit{A frog and a lizard}''. This strategy is simple and highly flexible.

\begin{figure}[h]
	\centering
	\setlength{\abovecaptionskip}{0pt}
	\setlength{\belowcaptionskip}{0pt}
	\includegraphics[trim=0 0 0 0bp, clip, width=0.9\linewidth]{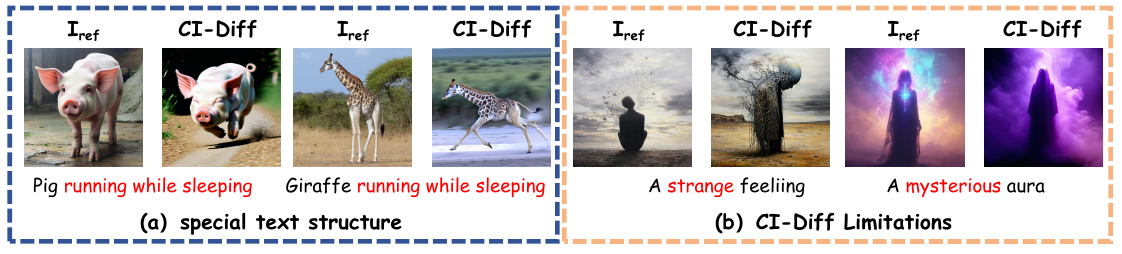}
	\caption{Success and failure cases.}
	\label{fig:fig2}
\end{figure}

\noindent \textbf{Discussion on limitations.} As stated in Sec.~\ref{sec:intro}, \textbf{CI-Diff} enhances the representation of rare attributes by disentangling them from rare text prompts, thereby enabling generalization to arbitrary sentence structures, as illustrated in Fig.~\ref{fig:fig2}(a). However, \textbf{CI-Diff} struggles to generate satisfactory images for abstract concepts, \textit{e.g.}, ``\textit{feeling}'' and ``\textit{aura}'' in Fig.~\ref{fig:fig2}(b), since \textbf{CI-Diff} primarily focuses on rare text prompts whose rarity arises from uncommon combinations of common adjective attributes and noun subjects.

\begin{figure*}[h]
	\centering
	\includegraphics[width=1\linewidth]{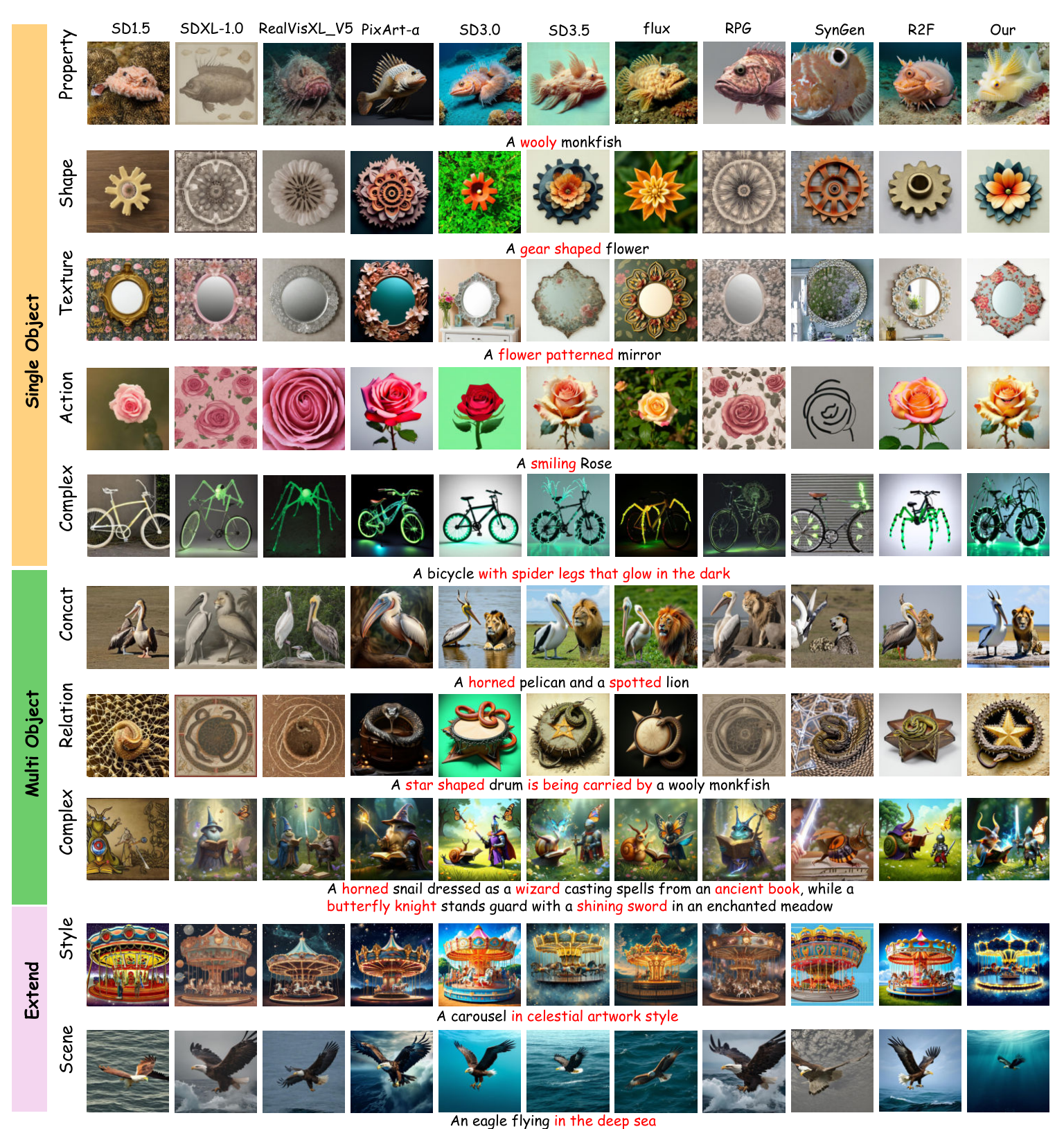}
	\caption{Qualitative comparison of \textbf{CI-Diff} with state-of-the-art diffusion baselines on RareBench. High-resolution visual generation results covering all original RareBench categories and two newly extended style and scene categories are displayed for comprehensive comparison.}
	\Description{Qualitative comparison showing our method outperforms baselines in structural integrity and rare attribute generation.}
	\label{fig:dingxingfenxi2}
\end{figure*}

\end{document}